\pdfoutput=1

\documentclass[11pt]{article}

\usepackage[final]{acl2023}

\usepackage{times}
\usepackage{latexsym}
\usepackage{makecell}
\usepackage[T1]{fontenc}

\usepackage[utf8]{inputenc}

\usepackage{microtype}

\usepackage{inconsolata}

\usepackage{graphicx}

\usepackage{array}
\usepackage{ragged2e}

\definecolor{clueGreen}{RGB}{0,130,0}
\definecolor{clueOrange}{RGB}{190,110,0}

\usepackage{listings} 

\usepackage[table]{xcolor}
\usepackage{colortbl}
\usepackage{tabularx} 
\usepackage{xcolor}
\usepackage{comment}
\usepackage{amssymb} 
\usepackage{multirow}
\usepackage{amsmath}
\usepackage{longtable}
\usepackage{booktabs}
\usepackage[most]{tcolorbox} 
\usepackage{pifont}
\newcommand{\xmark}{\textcolor{red}{\ding{55}}}
\newcolumntype{L}[1]{>{\RaggedRight\arraybackslash}p{#1}}
\newcolumntype{C}[1]{>{\Centering\arraybackslash}p{#1}}

\definecolor{darkgreen}{rgb}{0.0, 0.5, 0.0}

\title{MONETA: Multimodal Industry Classification through Geographic Information with Multi Agent Systems\thanks{The views expressed in this paper are personal views of the authors and do not necessarily reflect the views of Deutsche Bundesbank or the Eurosystem.}}

\author{
  \textbf{Arda Yüksel}$^{1, 2}$, 
  \textbf{Gabriel Thiem}$^{2, 3}$, 
  \textbf{Susanne Walter}$^{3}$, \\
  \textbf{Patrick Felka}$^{3}$, 
  \textbf{Gabriela Alves Werb}$^{3, 4}$, 
  \textbf{Ivan Habernal}$^{1, 5}$ \\
  \addlinespace[0.5em]
  $^1$Trustworthy Human Language Technologies
  $^2$Technical University of Darmstadt, Germany \\
  $^3$Deutsche Bundesbank
  $^4$Frankfurt University of Applied Sciences, Germany \\
  $^5$Research Center for Trustworthy Data Science and Security, Ruhr University Bochum, Germany \\
  \addlinespace[0.5em]
  \textbf{Correspondence:} \href{mailto:arda.yueksel@tu-darmstadt.de}{\texttt{arda.yueksel@tu-darmstadt.de}}
}

\begin{document}
\maketitle
\begin{abstract}
Industry classification schemes are integral parts of public and corporate databases as they classify businesses based on economic activity. Due to the size of the company registers, manual annotation is costly, and fine-tuning models with every update in industry classification schemes requires significant data collection. We replicate the manual expert verification by using existing or easily retrievable multimodal resources for industry classification. We present \texttt{MONETA}, the first multimodal industry classification benchmark with text (Website, Wikipedia, Wikidata) and geospatial sources (OpenStreetMap and satellite imagery). Our dataset enlists 1,000 businesses in Europe with 20 economic activity labels according to EU guidelines (NACE). Our training-free baseline reaches 62.10\% and 74.10\% with open and closed-source Multimodal Large Language Models (MLLM). We observe an increase of up to 22.80\% with the combination of multi-turn design, context enrichment, and classification explanations. 
We release \texttt{MONETA} and code\footnote{\url{https://github.com/trusthlt/Moneta}}.
\end{abstract}

\section{Introduction}

\begin{figure}[t]
  \centering
  \includegraphics[width=\linewidth]{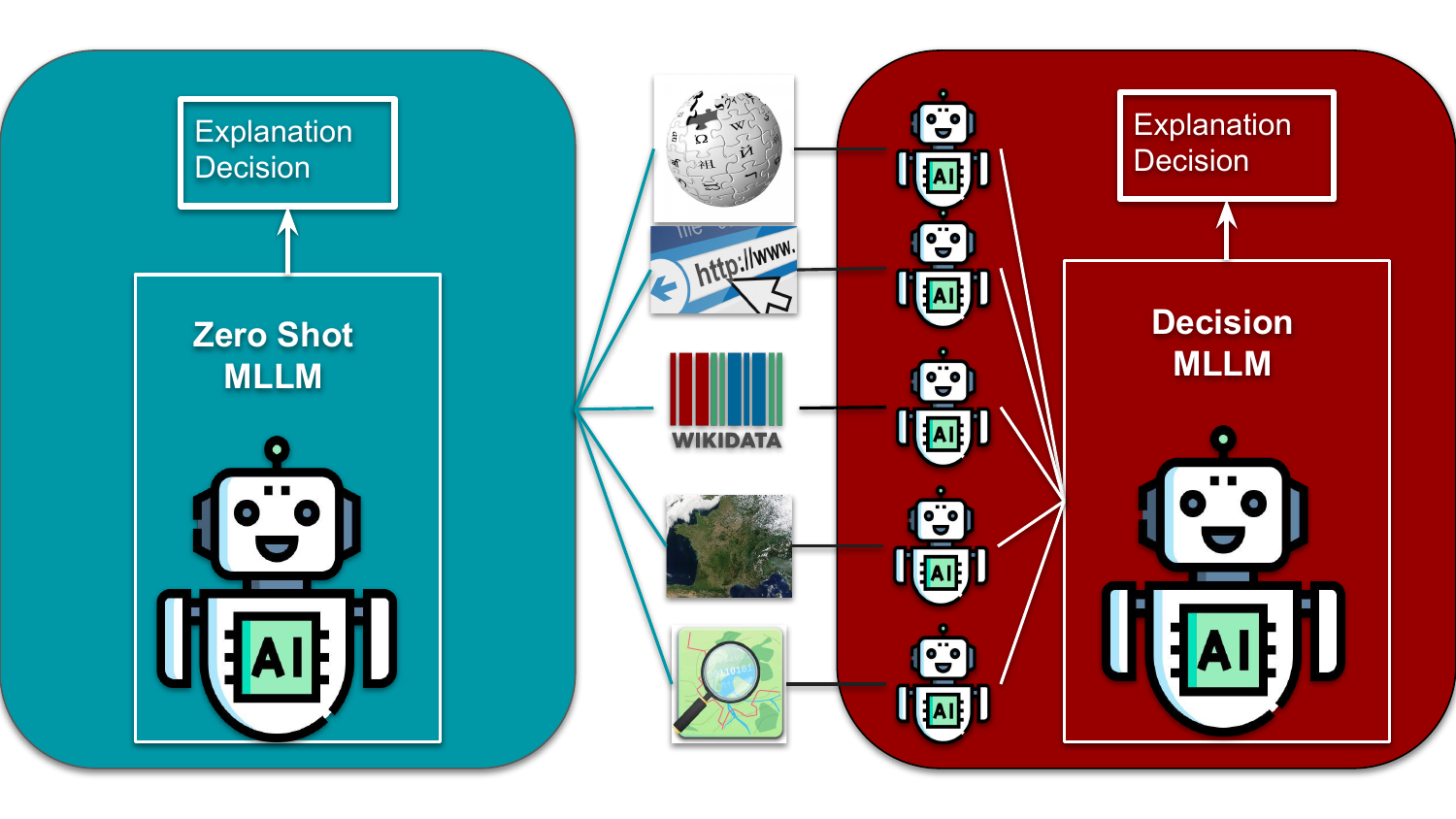}
  \caption{Zero-Shot vs Multi-Turn comparison for \texttt{MONETA}. 
  (1) Zero-shot pipeline (left): Available resources are forwarded into the industry classifier MLLM together. Explanations and classifications are obtained.
  (2) Multi-Turn pipeline (right): Each resource is processed by separate specialized agents. Intermediate \textit{clues} from these agents are processed by the decision-making agent, returning explanation and classification.}
  \label{fig:page 1-methdology}
\end{figure}

Geospatial finance \cite{Gopal2024GeospatialFinance} is an emergent and complex field that links financial and economic attributes to environmental and spatial resources. One of the important milestones in multimodal AI research, Multimodal Large Language Models (MLLM) such as Llava \cite{liu2023llava, liu2023improvedllava, liu2024llavanext}, InternVL \cite{chen2024expanding, wang2024mpo}, QwenVL
\cite{Qwen-VL, Qwen2-VL, Qwen2.5-VL}, GPT-5 \cite{openai2025gpt5} and Gemini 2.5 \cite{comanici2025gemini25pushingfrontier} can contribute to geospatial finance decision making tasks by processing visual geospatial data in addition to text documents. Traditionally, AI research has developed unimodal automatic industry classification based on global (ISIC \cite{un_international_2008}) and region-specific (NAICS \cite{ambler_introducing_1998}, NACE \cite{european_commission_nace_2008}) industry classification systems. National Statistical Business Registers classify enterprises by their primary economic activity using these classification schemes. The International Standard Industrial Classification (ISIC \cite{un_international_2008}) is widely used for economic statistics, including production, national accounts, employment, and population studies. 

Existing research on automatic industry classification from company recordings, financial reports, and websites \cite{kuhnemann_2020_nace_webtext, bechara_applying_2022, rizinski_company_2023, faria2023classifying, vamvourellis_company_2023, malashin_application_2024, guo_group_2025, dzuyo_linking_2025} has two main drawbacks. First, these methods rely solely on text, which is often unavailable for newly founded or small firms, whereas geospatial information from business registers can provide useful signals. Second, they fine-tune models that require large datasets and limit them to a single classification scheme. 

We propose, \texttt{MONETA}, a multimodal industry classification benchmark. On this task, Figure~\ref{fig:page 1-methdology}, we link text and geospatial resources to the economic activities using Zero-Shot and Multi-Turn pipelines.
Beyond the pure improvement of company registry information, further information extracted with our pipeline could be practically used for evaluating the sustainability of company assets (which is highly correlated with the classification of the primary economic activity) and their exposure to physical climate risks (wildfires, flooding). Supervisors and banks need this information to assess the vulnerability of the company assets that are financed through financial institutions.

\begin{table*}[t]
\centering
\footnotesize
\resizebox{\textwidth}{!}{%
\begin{tabular}{lccccc}
\toprule
\textbf{Dataset} & \textbf{Classes} & \textbf{Samples} & \textbf{Text Source} & \textbf{Image Source} & \textbf{Industry Scheme} \\
\midrule
\multicolumn{6}{c}{\textit{Remote Sensing Classification Benchmarks}} \\
\midrule
UC Merced \cite{yang_bag--visual-words_2010}            & 21  & 2,100     & \xmark  & Satellite         & \xmark \\
AID   \cite{xia_aid_2017}                & 30  & 10,000    & \xmark  & Satellite      & \xmark \\
AID++  \cite{jin_aid_2018}                & 46  & 400,000+  & \xmark  & Satellite  & \xmark \\
CLRS   \cite{li_clrs_2020}               & 25  & 15,000+   & \xmark  & Satellite       & \xmark \\
\midrule
\multicolumn{6}{c}{\textit{Industry Classification Benchmarks}} \\
\midrule
Dutch Businesses \cite{kuhnemann_2020_nace_webtext} & 111 & 40,796 & Websites & \xmark & NACE \\
GHAZAF  \cite{bechara_applying_2022}              & 56  & $\sim$6,500 & Survey text          & \xmark  & ISIC \\
SIRENE \cite{faria2023classifying} & 732 & $\sim$10 Million & Company Descriptions & \xmark & NACE \\
WRDS \cite{rizinski_company_2023}     & 11  & 34,338    & Company Descriptions & \xmark  & GICS \\
SEC 10K \cite{vamvourellis_company_2023}  & 11 (66)  & 2,590    & Company Descriptions & \xmark  & GICS \\
Industry  Websites \cite{jagric_ai_2024}   & 13  & 66,886    & Website & \xmark  & Custom \\
Economic Activity Records \cite{malashin_application_2024} & 20 (88) & $\sim$20 Million &  Company Descriptions & \xmark & NACE \\
SEC EDGAR  \cite{dzuyo_linking_2025}      & 8   & 9,582     & Financial reports     & \xmark  & SIC \\
ExioNAICS \cite{guo_group_2025}             & 20 (1,114)  & 20,850    & Descriptions + emissions & \xmark  & NAICS \\

\midrule
\multicolumn{6}{c}{\textit{Our Dataset}} \\
\midrule
\textbf{\texttt{MONETA} (OURS)}         & \textbf{20} & \textbf{1,000} & \textbf{Website + Wikipedia + Wikidata} & \textbf{Satellite + OSM} & \textbf{NACE} \\
\bottomrule
\end{tabular}%
}
\caption{Comparison of datasets across remote sensing and industry classification tasks. \xmark\ indicates the absence of that modality or label scheme.}
\label{tab:benchmark_comparison}
\end{table*}

We connect economic activities to spatial extent to answer following research questions:
\begin{itemize}
    \item RQ-1: Can MLLMs use geospatial information as well as text for industry classification?
    \item RQ-2: Which configuration (classification explanations, context enrichment, multi-agent) is more helpful?
    \item RQ-3: How can we quantify intermediate agent performance with respect to the final prediction and the ground truth labels?
\end{itemize}
In this study, we propose a novel task: \textbf{Multimodal Industry Classification with Geospatial Information} and introduce:
\begin{itemize}
    \item \textbf{\texttt{MONETA}: Multimodal Industry Classification Benchmark} for 1,000 European businesses in 20 NACE \cite{european_commission_nace_2008} sections. We provide two visual resources (OpenStreetMap (OSM) and Satellite) and at least one text resource (Wikidata, Wikipedia, and website) per entry. 
    \item \textbf{Multimodal Industry Classification}: An expert domain multimodal AI task rooted in geospatial finance. We propose Zero Shot and Multi-Turn (Multi-Agent) approaches supporting various multimodal resources, output configurations, and prompting strategies.
    \item \textbf{Novel Intermediate Agent Evaluation:} We provide quantitative measures for final inference certainty. Also, we propose a novel keyword-based strategy to analyze intermediate agent performance with respect to ground truth and decision-making agent prediction.
\end{itemize}

\section{Related Work}

\subsection{Industry Classification}
Industry classification dates back to the 1930s with the Standard Industrial Classification (SIC), supporting market and sustainability analysis \cite{ambrois_2023_EIF,croce_cleantech_2024}. Automated business classification remains an active research area \cite{kuhnemann_2020_nace_webtext, faria2023classifying}. To reflect emerging sectors, multiple schemas have been introduced, including GICS by MSCI and S\&P,\footnote{https://www.msci.com/indexes/index-resources/gics}
 and ISIC by the United Nations \cite{un_international_2008}, with regional variants such as NAICS \cite{ambler_introducing_1998} and NACE \cite{european_commission_nace_2008}. The European Union uses NACE, a hierarchical ISIC-based scheme with 21 sections (A–U) representing major economic activities (e.g., \textit{C: Manufacturing}), followed by 88 divisions, 272 groups, and 514 classes.
 


Industry classification provides insight into market and sustainability analysis \cite{ambrois_2023_EIF,croce_cleantech_2024}. \citet{jagric_ai_2024} emphasizes that industry codes affect assessments of market competition, regulatory decisions, and market research outcomes. \citet{kuhnemann_2020_nace_webtext} note that determining a firm’s main industry activity is a critical yet challenging task in statistical business registers and typically relies on expert judgment and consultation. \citet{werb_geospatial_2024} further report that existing approaches remain largely manual due to data inconsistencies and the heterogeneous, multimodal nature of company master data, even at the scale of national institutions such as the German Federal Bank. Despite expert involvement and the importance of the task, misclassifications are common.

For automatic industry classification, \citet{kuhnemann_2020_nace_webtext} used websites for NACE classes. 
\citet{rizinski_company_2023}, \citet{faria2023classifying} and \citet{vamvourellis_company_2023} used company descriptions to classify industries. Fine-tuning transformers and adapters is a common approach \cite{bechara_applying_2022, jagric_ai_2024, guo_group_2025, dzuyo_linking_2025} for coarse and fine-grained industry classification. \citet{malashin_application_2024} employed a genetic algorithm approach for hyperparameter tuning for NACE's divisions.

Many of the studies above, except \citet{rizinski_company_2023}, relied on fine-tuning, which requires extensive data collection and annotation and makes the model unusable for other schemas. Furthermore, all the studies incorporated unimodal text sources, such as financial statements, which may not be available for newly-founded companies. \citet{werb_geospatial_2024} argue that economic activity analysis can be enriched with other sources such as satellite imagery, which is the research gap this study covers.

\subsection{Geospatial Understanding}
Geospatial AI (GeoAI) methodologies cover a variety of tasks such as Geolocation, \cite{song_geolocation_2025, mendes_granular_2024}, Geocoding \cite{nakatani_text_2025}, Remote Sensing \cite{tao_graph-based_2025}, Question Answering and Fact Verification \cite{norouzi_knowledge-enhanced_2025, anderson_measuring_2025, khan_debunking_2024},  Geospatial Foundation Model and Agents \cite{mansourian_chatgeoai_2024, xu_rs-agent_2024}.

Remote sensing extracts geospatial features from sources such as satellite imagery, street views, and OpenStreetMap (OSM\footnote{https://www.openstreetmap.org/}
), which can link to external resources like Wikipedia, Wikidata, and websites. The AI community has developed many remote sensing datasets, including UC-MED \cite{yang_bag--visual-words_2010} for land-use classification, AID and AID++ \cite{xia_aid_2017, jin_aid_2018} for aerial scene understanding, and CLRS \cite{li_clrs_2020} for continual learning.

Recent GeoAI research fuses multimodal data to infer economic attributes in urban contexts; however, these approaches typically yield regional proxies or area-based estimates rather than entity-specific financial data \cite{tao_graph-based_2025, chen_fusing_2025}. For instance, \citet{yang_poverty_2024} linked satellite imagery (2015–2023) to socio-economic deprivation to provide a neighborhood-scale assessment. At the core of this field is the spatio-temporal dimension, which enables diverse applications ranging from monitoring human-nature relations \cite{yuan_fusu_2024} to analyzing climate change impacts and extreme weather \cite{chen_2024_terra, shams_2024_idee}. Building on these foundations, \citet{li-etal-2025-llms} bridges these spatio-temporal signals with Large Language Models (LLMs) to transform Health \& Public Services (H\&PS). In their work, they specifically address the challenge of hallucinations by evaluating how models navigate spatial and temporal signals. Given the risks of misinformation, \citet{li-etal-2025-llms} demonstrate that it is essential to enrich spatial elements with additional context in high-stakes domains.

Despite these advances, prior works do not address entity-level industry classification, as reflected in Table~\ref{tab:benchmark_comparison}. Our work is the first to study the suitability of geospatial resources for the industry classification of individual businesses. For robust and traceable analysis, we combine spatial sources with a plethora of textual information. 

\begin{figure*}[!h]
    \centering    \includegraphics[width=\textwidth]{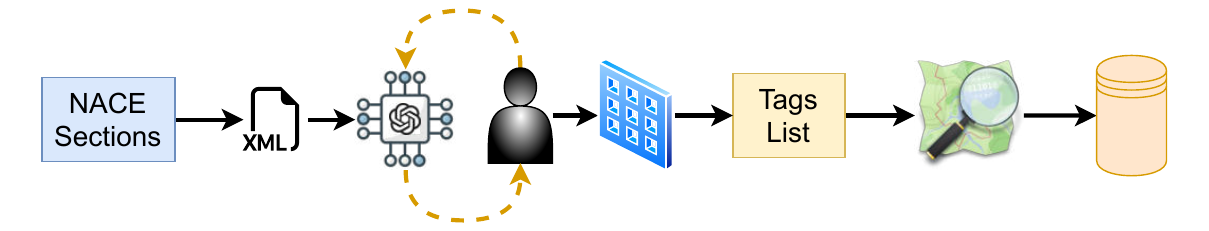}
    \caption{Overview of the dataset preparation process. (1) NACE section XMLs are initially converted to the OSM tags and manually checked by the authors. (2) We added custom filters for data quality and queried Europe OSM data with tag list. (3) Samples are grouped by NACE codes to form the gold dataset. } 
    \label{fig:dataset_construction}
\end{figure*}

\section{\texttt{MONETA}}
We introduce \texttt{MONETA}, a novel multimodal benchmark for industry classification based on EU Guidelines (NACE) for European businesses. In this section, we explain our mapping and dataset. 

\noindent\textbf{Mapping: } Due to a lack of direct mapping connecting OSM and NACE sections, we generated a \textbf{novel NACE to OSM} mapping using the methodology shown in Figure~\ref{fig:dataset_construction}. In addition to the code and the dataset, we also release the proposed mapping for further research.

We first used Gemini to generate OSM tags for NACE sections from official guidelines in RDF/XML. Because this mapping was error-prone, we introduced a human-in-the-loop process and refined the annotations using GPT and Gemini. This resulted in a validated list of OSM tags per NACE section. We then applied data-quality filters (name, address, and external links such as website, Wikipedia, and Wikidata). Finally, we queried OSM, grouped entries by NACE section, and obtained the gold dataset. Additional details and examples are given in Appendix~\ref{sec:appendix_nace_osm}.

\noindent\textbf{Gold Dataset:} We sampled 50 entries per NACE section (A–U, excluding T) and formed the first multimodal industry classification benchmark with 1,000 businesses. Using bounding boxes, we computed dynamic zoom levels and retrieved satellite imagery via the ESRI REST API.\footnote{https://developers.arcgis.com/rest/static-basemap-tiles/} ESRI and OSM services return static tiles for given coordinates and zoom levels; concatenating these tiles yields aligned OSM and satellite images of the same area. None of the external resources in \texttt{MONETA} explicitly mentions the NACE section in their context. Additional properties are available in Appendix \ref{appendix:data_properties}.


\section{Experiments}
Multimodal industry classification task tests MLLM using various resources to predict economic activities in two pipelines: Zero-Shot and Multi-Turn. Zero-Shot detects NACE section in single inference using multimodal inputs. Multi-turn has clue extracting agents for each input type, and a decision-making agent processes these clues.

We have several experiment dimensions for model selection, prompting strategies, input configurations, and output structures. Using frequency vectors in the clue analysis stage, we quantify intermediate agent effectiveness and correctness. We propose new metric to analyze model uncertainty.

\subsection{Pipeline}
In this study, we tested two adaptable and training-free pipelines to accommodate future changes in classification schemes: Zero-Shot and Multi-Turn. 

\noindent \textbf{Zero-Shot: }We provide inputs with various configurations and instruct MLLM to utilize them to classify entities based on NACE sections. 

\noindent \textbf{Multi-Turn: }Multi-turn has two stages: Clue Extraction and Decision Making. The clue extraction contains agents designed to generate \textit{clues} up to the number of inputs (e.g., OSM). Decision-making agent uses intermediate agent responses, \textit{clues}, and the entity name to choose NACE sections.

\begin{figure*}[!h]
    \centering
    \includegraphics[width=\linewidth]{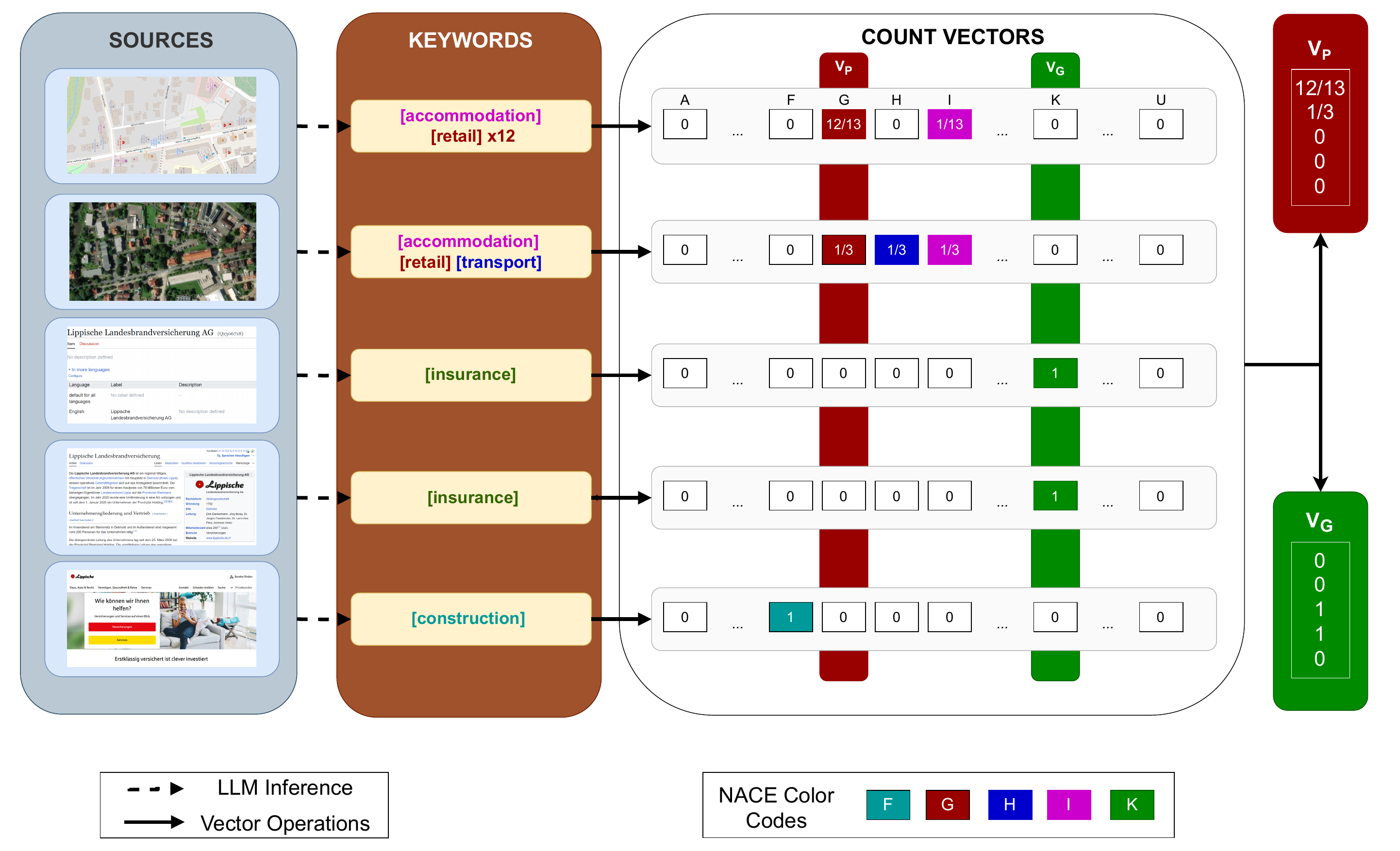}
    \caption{Example with ground truth NACE sections K and prediction G. Clues and predictions are obtained via InternVL-3 8B. 
    Clue Analysis Methodology: (1) Sources are forwarded into separate MLLM Agents for clue extraction. (2) Keywords based on predefined Economic Activities are extracted and grouped into NACE sections. (3) For each resource (OSM, Satellite, Wikidata, Wikipedia and Website), normalized count vectors are formed. The grouped keywords are first placed to a vector with 21 dimensions (number of sections) and then divided by the total number of keywords for inference. (4) Ground-Truth and Final Prediction NACE sections are used to form $V_G$ and $V_P$. Both of these vectors have the dimensions of 5 (number of inputs). }
    \label{fig:count_vector}
\end{figure*}

\subsection{Experiment Dimensions}

\textbf{Models: }We selected open and closed-source models: InternVL 2.5 (1B, 4B, 38B) and InternVL 3 (8B, 14B, 38B, 78B) \cite{chen2024expanding, wang2024mpo}, Llava v1.6 (7B, 13B, 34B) \cite{liu2023llava, liu2023improvedllava, liu2024llavanext}
and QwenVL 2.5 (7B, 32B, 72B) \cite{Qwen-VL, Qwen2-VL, Qwen2.5-VL}, Gemini 2.5 \cite{comanici2025gemini25pushingfrontier}, GPT 5 - Mini, and GPT 5.1 \cite{openai2025gpt5}. The details of frameworks and infrastructure are given in Appendix \ref{sec:implementation}.

\noindent\textbf{Prompt Templates:} We have two decision-making prompt templates: \texttt{Simple} and \texttt{Extended}. In both prompts, we include NACE section codes and titles. \texttt{Extended} prompt has section summaries from official guidelines with description, their content, and exclusions. Templates and contents are given in the Appendix \ref{sec:prompts}.

\noindent\textbf{Input Configurations:} In all the classification experiments, we included name of the entity in the context. In addition to this, we tested with single inputs for satellite image or external resource. As having OSM content implies at least one other resource, we did not use OSM as a single resource. We also tested combination of inputs: Satellite + OSM, Satellite + External and All. 

\noindent\textbf{Output Structure:} We support two output structures for free text generation. In \textit{Text} output, we instructed MLLM to generate single token answers from NACE sections or \texttt{UNK} if uncertain. To analyze the effect of classification explanations, in another output structure, we instructed MLLM to return JSON with the explanation and the decision.

\subsection{Clue Analysis}\label{sec:method_clue}

In our multi-turn pipeline, clue agents are instructed to process specific input types to generate free-form texts with keywords, Table~\ref{tab:appendix_nace_content}, describing economic activities. For example: [retail] for section G (Wholesale and Retail Trade; Repair of Motor Vehicles and Motorcycles). In case of no evidence, agent returns No Economic Activity Found.

Through keywords, we can analyze the free-form text as shown in Figure~\ref{fig:count_vector}. For this example, prediction is G while the correct result is K. From the satellite image, we found evidence [accommodation] (I), [retail] (G), [transport] (H). From Wikidata, we found a single keyword [insurance] (K).

Upon grouping keywords by sections, we obtain normalized keyword counts. We refer to this scaled column vector as the frequency vector,  $v_{i, c}$. In the example, the satellite frequency vector contains $1/3$ for sections G, H and I, and the Wikidata frequency vector has $1$ for section K. The remaining values will be 0. If an agent fails to identify economic activity, we would have a vector of 0s.

We can use these frequency vectors to emphasize on ground truth label $g$, and the final prediction $p$. By selecting these indices, we can formulate ground truth and final prediction frequency vectors: 

\begin{align}
\begin{array}{l}
\text{Ground Truth: } v_{g_i}[i] \\
\text{Prediction: } v_{p_i}[i]
\end{array}
&=
\begin{bmatrix}
v_{i,c=\mathrm{OSM}}[g_i \mid p_i] \\
v_{i,c=\mathrm{Satellite}}[g_i \mid p_i] \\
v_{i,c=\mathrm{Wikidata}}[g_i \mid p_i] \\
v_{i,c=\mathrm{Wikipedia}}[g_i \mid p_i] \\
v_{i,c=\mathrm{Website}}[g_i \mid p_i]
\end{bmatrix}
\end{align}


In the example Figure \ref{fig:count_vector}, we use index K for
the ground truth vector. For OSM, $v_{\mathrm{OSM}}[K]$, satellite $v_{\mathrm{Satellite}}[K]$, and website, $v_{\mathrm{Website}}[K]$, results are 0. For wikidata $v_{\mathrm{Wikidata}}[K]$ and Wikipedia $v_{\mathrm{Wikipedia}}[K]$ results are 1. We form the ground truth vector, $v_{g} = [0; 0; 1; 1; 0]$. By changing the index with prediction label, G, we can retrieve the prediction vector, $v_{p} = [12/13; 1/3; 0; 0; 0]$. 

\subsection{Metrics}
In this study, in addition to \textit{Accuracy}, we introduce \textit{Unknown Ratio (UR)}. It is calculated using number of predictions with \texttt{UNK}, $U$,  response over total number of inferences, $I$:
\begin{align}
    \text{Unknown Ratio (UR)} &= U / I
\end{align}

To evaluate clue extraction, in our multi-turn pipeline, we propose additional metrics:
\begin{itemize}
    \item \textbf{Correctness: }Measures relatedness of clues to ground truth labels. It is the sum of all of ground truth vectors, $v_g[clue=c]$, divided by the number of inferences for the input, $I_c$. 
    \begin{align}
        \text{Correctness}_c &= \frac{\sum_{i}^{I_c}v_{g_i} [i, c]}{I_c}
    \end{align}
    For example, using $I_c = 1$ (due to single inference), we can find the correctness of Wikidata and Wikipedia, as $V_{G=K}[c=Wikidata | Wikipedia] = 1$. Other inputs have 0 in $V_G$, so they have 0 correctness.
    \item \textbf{Effectiveness: }Measures how clues affect the final predictions.  It is the sum of all of the prediction vectors, $v_p[clue=c]$, divided by the number of inferences for the resource, $I_c$.
    \begin{align}
        \text{Effectiveness}_c &= \frac{\sum_{i}^{I_c}v_{p_i} [i, c]}{I_c}
    \end{align}
    In the example, only satellite and OSM have non-zero values in $V_P$, which are $1/3$ and $12/13$. Since we have one inference, their effectiveness will be $1/3$ and $12/13$.      
\end{itemize}

\section{Results}

\begin{table*}[]
\centering
\resizebox{\textwidth}{!}{%
\begin{tabular}{ll|rrrrrr}
\toprule
 \textbf{Model} &
  \textbf{Size (B)} &
  None &
  \textbf{\textcolor{orange}{Satellite}} &
  \textbf{External} &
  \textbf{\textcolor{orange}{Satellite} + \textcolor{orange}{OSM}} &
  \textbf{\textcolor{orange}{Satellite} + External} &
  \textbf{All} \\
\midrule
\multicolumn{8}{c}{Open Source Models} \\
\midrule
\multirow{3}{*}{\textbf{InternVL 2.5}} & 1  & \textbf{4.20}  & 2.20          & 1.00           & 2.50  & 0.90           & 0.20           \\
& 4  & 8.70           & 6.60           & 11.10          & 4.70  & \textbf{13.50} & 6.40           \\
& 38 & 46.30          & 49.80          & 58.40          & 51.40 & \textbf{61.40} & 60.10          \\\midrule
\multirow{4}{*}{\textbf{InternVL 3}}   & 8  & 43.60          & 34.90          & \textbf{48.10} & 30.10 & 46.90          & 41.70          \\
& 14 & 45.00          & 49.30          & \textbf{56.10} & 48.30 & 55.60          & 53.00          \\
& 38 & 44.60          & 49.20          & 58.60          & 49.00 & \textbf{59.80} & 58.30          \\
& 78 & 43.40          & 47.80          & 60.40          & 46.10 & \textbf{62.10} & 58.80          \\\midrule
\multirow{3}{*}{\textbf{Llava 1.6}}    & 7  & 1.50           & \textbf{2.20}  & \xmark           & \xmark  & \xmark           & \xmark           \\
& 13 & \textbf{13.10} & 12.30          & \xmark           & \xmark  & \xmark           & \xmark           \\
& 34 & 1.20           & \textbf{16.60} & \xmark           & \xmark  & \xmark           & \xmark           \\\midrule
\multirow{3}{*}{\textbf{QwenVL 2.5}}   & 7  & 19.80          & 19.10          & 22.10          & 17.60 & 21.80          & \textbf{23.10} \\
& 32 & 45.30          & 48.60          & \textbf{57.50} & 46.30 & 57.00          & 56.40          \\
& 72 & 46.20          & 43.90          & 56.90          & 45.50 & 59.30          & \textbf{60.50}\\
\midrule
 \multicolumn{8}{c}{Closed Source Models} \\
\midrule
 \multicolumn{2}{l|}{Gemini 2.5 Flash} & 58.40 & 63.50 & 71.00 & 66.80 & 73.80 & 72.40 \\
 \multicolumn{2}{l|}{GPT 5 Mini} & \textbf{62.00} & \textbf{66.80} & \textbf{71.90} & \textbf{68.90} & \textbf{74.10} & \textbf{73.30} \\
 \multicolumn{2}{l|}{GPT 5.1} & 57.80 & 59.60 & 69.10 & 63.40 & 70.00 & 70.20 \\
\bottomrule
\end{tabular}%
}
\caption{Baseline (\textit{Zero-Shot} pipeline, \textit{Simple} prompt, \textit{Text} output) accuracy for NACE industry classification. Columns after model and size denote input configurations. \textcolor{orange}{Image} inputs are highlighted. \textbf{Bold} indicates the best performance for the model and size pair for the open-source model, and the best performance among closed-source models. GPT 5 Mini and InternVL 3-78B are the best-performing closed and open source models.}
\label{tab:nace_baseline}
\end{table*}

\subsection{Baseline}

We demonstrated the baseline results in Table~\ref{tab:nace_baseline} using the Zero-Shot pipeline, \texttt{Simple} prompt, and \texttt{Text} output. The name of the entity is given for every input configuration.

InternVL 3-78B and GPT 5-Mini achieved the highest performance for open and closed-source MLLMs. InternVL3-14B is the best-performing small ($\leq14B$) model. Due to its limited context window and weaker performance, we exclude LLaVA v1.6 from the remaining experiments. The difficulty of the task is apparent as even 70B+ models failed to reach \textbf{65\%} accuracy. Also, the best open source performance is on par with the best model's name-only performance, which indicates the gap between open and closed source MLLMs.

\noindent \textbf{RQ-1: Can MLLMs use geospatial information as well as text for industry classification?}

\begin{figure}[!h]
    \centering
    \includegraphics[width=\linewidth]{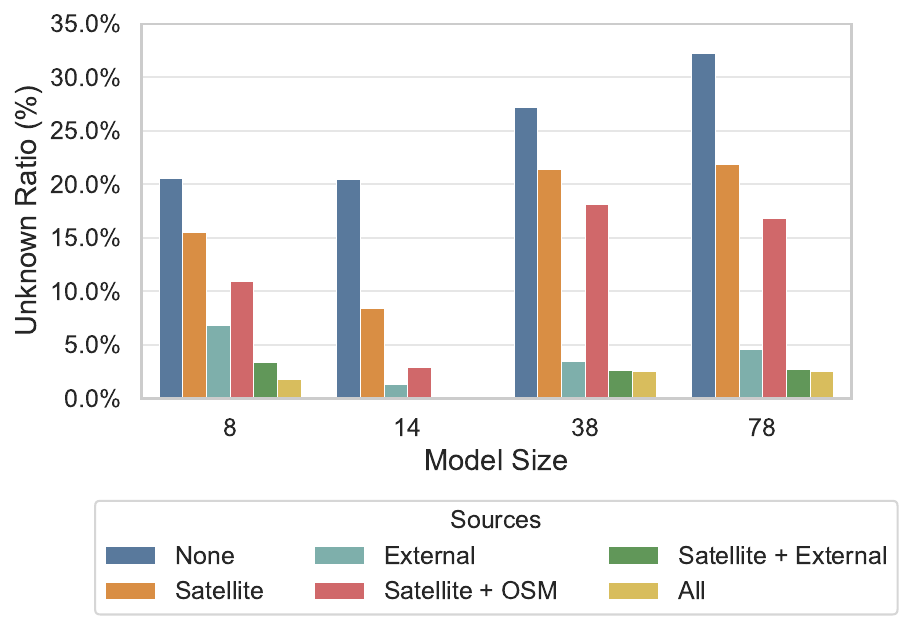}
    \caption{Baseline unknown ratio of InternVL 3 for input configurations. Unknown ratio corresponds to \texttt{UNK} \textcolor{blue}{(uncertain)} responses over all inferences.}
    \label{fig:baseline_accuracy_ur_vs_input}
\end{figure}

\noindent To quantify uncertainty of the InternVL 3 responses, we listed unknown ratios in Figure~\ref{fig:baseline_accuracy_ur_vs_input}. Our two assumptions were that adding more inputs would increase performance and reduce uncertainty. However, we identified that accuracy increase is not guaranteed, especially with smaller models. Providing additional inputs yields accuracy gains of at most \textbf{20\%}, making the entity name the strongest predictive signal. Furthermore, model performance is the best when external resources are given in context compared to geospatial resources. 

Unlike accuracy, the unknown ratio reveals that the name alone is not enough for a robust prediction. We also noted that the uncertainty decreases significantly more when text information is provided compared to visual information. However, one must note that the image inputs reveal neighborhood information while the external inputs map directly to the entity.

\begin{table*}[]
\centering
\resizebox{\textwidth}{!}{%
\begin{tabular}{lc|r|rrrrr}
\toprule
    \textbf{Model} &
    \textbf{Size} &
    Baseline &
   Explanation&
   Extended Prompt &
   Multi-Turn &
  Extended Prompt + &
  Mixture \\
  \midrule
  \multicolumn{8}{c}{Open Source Models} \\
  \midrule
\multirow{2}{*}{InternVL 2.5} &
  4B &
  6.40 &
  23.10 (\textcolor{darkgreen}{16.70}) &
  8.30 (\textcolor{darkgreen}{1.90}) &
  22.00 (\textcolor{darkgreen}{15.60}) &
  10.60 (\textcolor{darkgreen}{4.20}) &
  \textbf{29.20 (\textcolor{darkgreen}{22.80})} \\
 &
  38B &
  60.10 &
  61.80 (\textcolor{darkgreen}{1.70}) &
  64.20 (\textcolor{darkgreen}{4.10}) &
  58.20 (\textcolor{red}{-1.90}) &
  \textbf{65.00 (\textcolor{darkgreen}{4.90})} &
  60.20 (\textcolor{darkgreen}{0.10}) \\
\multirow{2}{*}{InternVL 3} &
  8B &
  41.70 &
  42.30 (\textcolor{darkgreen}{0.60}) &
  36.90 (\textcolor{red}{-4.80}) &
  \textbf{49.80 (\textcolor{darkgreen}{8.10})} &
  38.00 (\textcolor{red}{-3.70}) &
  45.70 (\textcolor{darkgreen}{4.00}) \\
 &
  38B &
  58.30 &
  59.80 (\textcolor{darkgreen}{1.50}) &
  61.30 (\textcolor{darkgreen}{3.00}) &
  61.60 (\textcolor{darkgreen}{3.30}) &
  \textbf{64.10 (\textcolor{darkgreen}{5.80})} &
  62.60 (\textcolor{darkgreen}{4.30}) \\
\multirow{2}{*}{QwenVL 2.5} &
  7B &
  23.10 &
  30.50 (\textcolor{darkgreen}{7.40}) &
  27.30 (\textcolor{darkgreen}{4.20}) &
  38.90 (\textcolor{darkgreen}{15.80}) &
  31.00 (\textcolor{darkgreen}{7.90}) &
  \textbf{45.90 (\textcolor{darkgreen}{22.80})} \\
 &
  32B &
  56.40 &
  60.00 (\textcolor{darkgreen}{3.60}) &
  60.40 (\textcolor{darkgreen}{4.00}) &
  55.70 (\textcolor{red}{-0.70}) &
  \textbf{65.40 (\textcolor{darkgreen}{9.00})} &
  62.00 (\textcolor{darkgreen}{5.60})\\
  \midrule
  \multicolumn{8}{c}{Closed Source Models} \\
  \midrule
   \multicolumn{2}{l|}{Gemini 2.5 Flash} & 72.40 & 72.50 (\textcolor{darkgreen}{0.10}) & \textbf{74.30 (\textcolor{darkgreen}{1.90})}  & 71.20 (\textcolor{red}{-1.20}) & 74.00 (\textcolor{darkgreen}{1.60}) & 72.70 (\textcolor{darkgreen}{0.30})\\
   \multicolumn{2}{l|}{GPT 5 Mini} & 73.30 & 74.00 (\textcolor{darkgreen}{0.70}) & \textbf{74.70 (\textcolor{darkgreen}{1.40})} & 74.30 (\textcolor{darkgreen}{1.00}) & 72.90 (\textcolor{red}{-0.40}) & 74.20 (\textcolor{darkgreen}{0.90})\\
   \multicolumn{2}{l|}{GPT 5.1} & 70.20 & 69.40 (\textcolor{red}{-0.80}) & 69.70 (\textcolor{red}{-0.50}) & 69.00 (\textcolor{red}{-1.20}) & 68.50 (\textcolor{red}{-1.70}) & \textbf{70.40 (\textcolor{darkgreen}{0.20})} \\
  \bottomrule
\end{tabular}%
}
\caption{Selected model accuracies with: Explanations, prompt context enrichment, multi-turn pipeline. Extended Prompt + is the combination of the extended prompt with explanations. Mixture is the combined setting with all the advancements. In these results, MLLMs used all inputs. The best result for a given model and size is shown in \textbf{bold}.}
\label{tab:nace_combined_with_diff_selected}
\end{table*}

\subsection{Configurations}
In our experiment setup, we allowed customization for several dimensions: output structure, prompt template, and pipeline. For open-source models, we selected one small ($\leq8$) and one large ($\geq30B$) model for InternVL 2.5 and 3 and QwenVL 2.5. The accuracy results are shown in Table \ref{tab:nace_combined_with_diff_selected}. For these experiments, we used all the inputs available.

\noindent \textbf{RQ-2: Which configuration (classification explanations, context enrichment, multi-agent) is more helpful for the task?}

\noindent The smaller models perform better with the Multi-Turn pipeline. Its combination with prompt enrichment and explanations gives more than 20\% boost InternVL 2.5-4B and QwenVL 2.5-7B, while InternVL 3-8B reaches almost 50\% with only Multi-Turn. For larger ($\geq30B$) and proprietary models, we obtained the best performances without multi-turn. The best performing smaller model is also the most recent model, InternVL 3-8B. 

\subsection{Clue Analysis}

We extracted frequency vectors from keywords for intermediate agent \textit{clues}. From frequency vectors, we can measure how \textit{effective} each input is to the final prediction and how \textit{correct} these responses are with respect to ground truth. We demonstrated InternVL 3 (8B and 38B) results for the mixture configuration (multi-turn with extended prompt and classification explanation) in Figure~\ref{fig:internvl3_ce_plot}. Other model results are available in Appendix Table~\ref{tab:clue_correctness_effectiveness}.

\begin{figure}[!h]
    \centering
    \includegraphics[width=\linewidth]{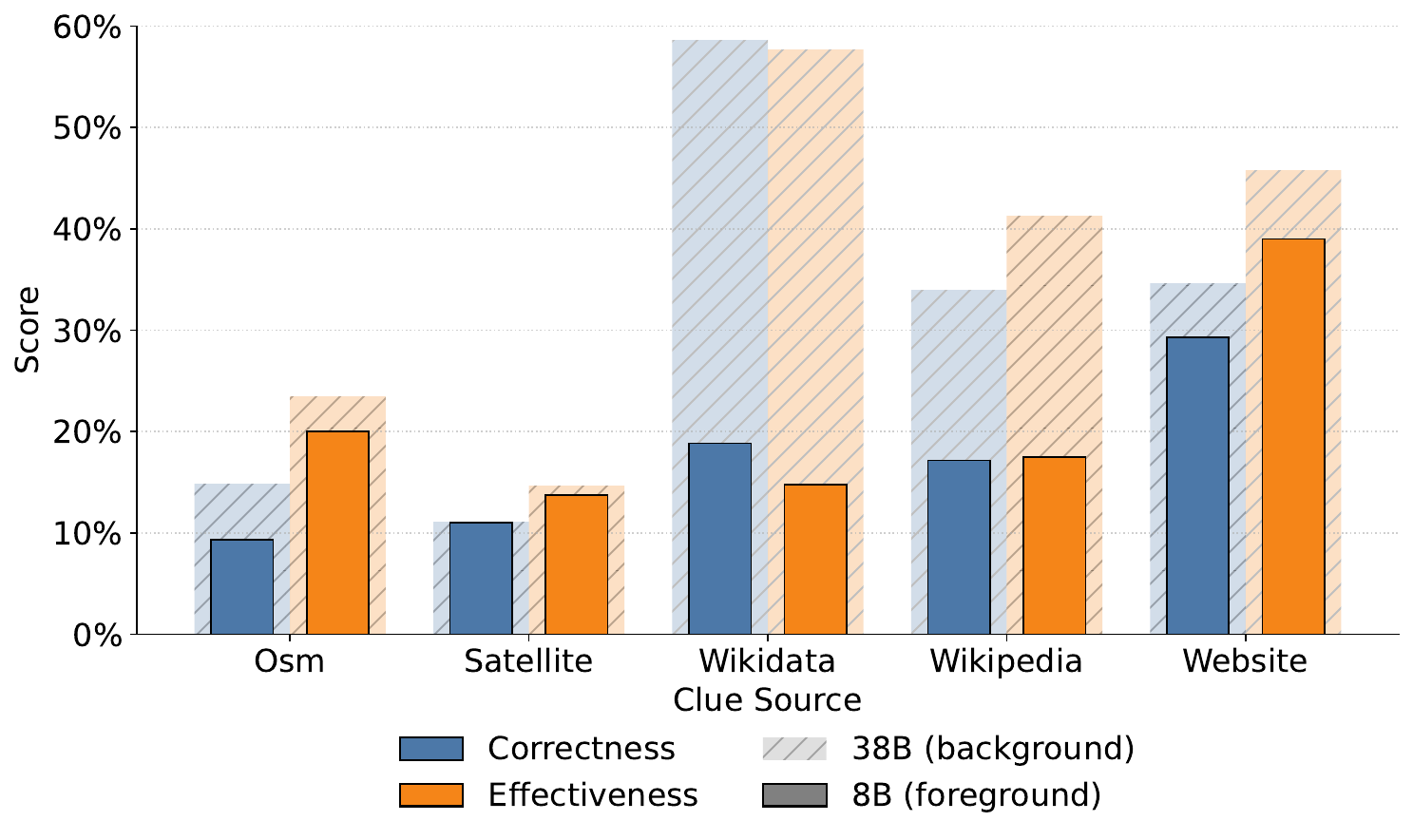}
    \caption{InternVL 3 (8B and 38B) correctness and effectiveness scores for each input. In these experiments, multi-turn pipeline with extended prompt and classification explanations is used.}
    \label{fig:internvl3_ce_plot}
\end{figure}

\noindent \textbf{RQ-3: How can we quantify intermediate agent performance with respect to the final prediction and the ground truth labels?}

\noindent Both models generate more truthful clues from text sources, especially Wikidata and websites. Except for websites, the smaller model fails to generate useful and effective clues from most sources. For the larger model, Wikidata appears to be the best resource. For image inputs, results do not improve with scale. This indicates the difficulty of clue extraction from geospatial information. 

In all experiments, text input clues correlate more with the ground truth and are more effective for final prediction. This is expected because of two reasons: (1) Our text content is mostly present in or similar to the pretraining data, (2) the used models are not adapted to remote sensing.

\subsection{Ablations}
\textbf{Qualitative Results:} In Table~\ref{tab:qualitative_moneta_example}, we selected examples from \texttt{MONETA} containing satellite images and websites (translated and summarized). In these examples, the generated clues contradict each other. While websites are often the most effective source, they may emphasize sales-related information, introducing a bias toward NACE Section G (Wholesale and Retail Trade). When websites are absent or less informative, satellite imagery can instead enable correct identification of the industry.

However, as the quantitative results show for the second example, visual cues can also be misleading. For a robust and accurate industry classification, both text clues and visual clues should be used. 

\begin{table}[!t]
\scriptsize
\setlength{\tabcolsep}{6pt}
\renewcommand{\arraystretch}{1.1}
\centering

\begin{tabular}{p{0.45\linewidth} p{0.45\linewidth}}
\toprule

\textbf{Example 1} & \textbf{Example 2} \\[4pt]

\textbf{Inputs} & \textbf{Inputs} \\

\begin{minipage}[t]{\linewidth}
\includegraphics[width=\linewidth]{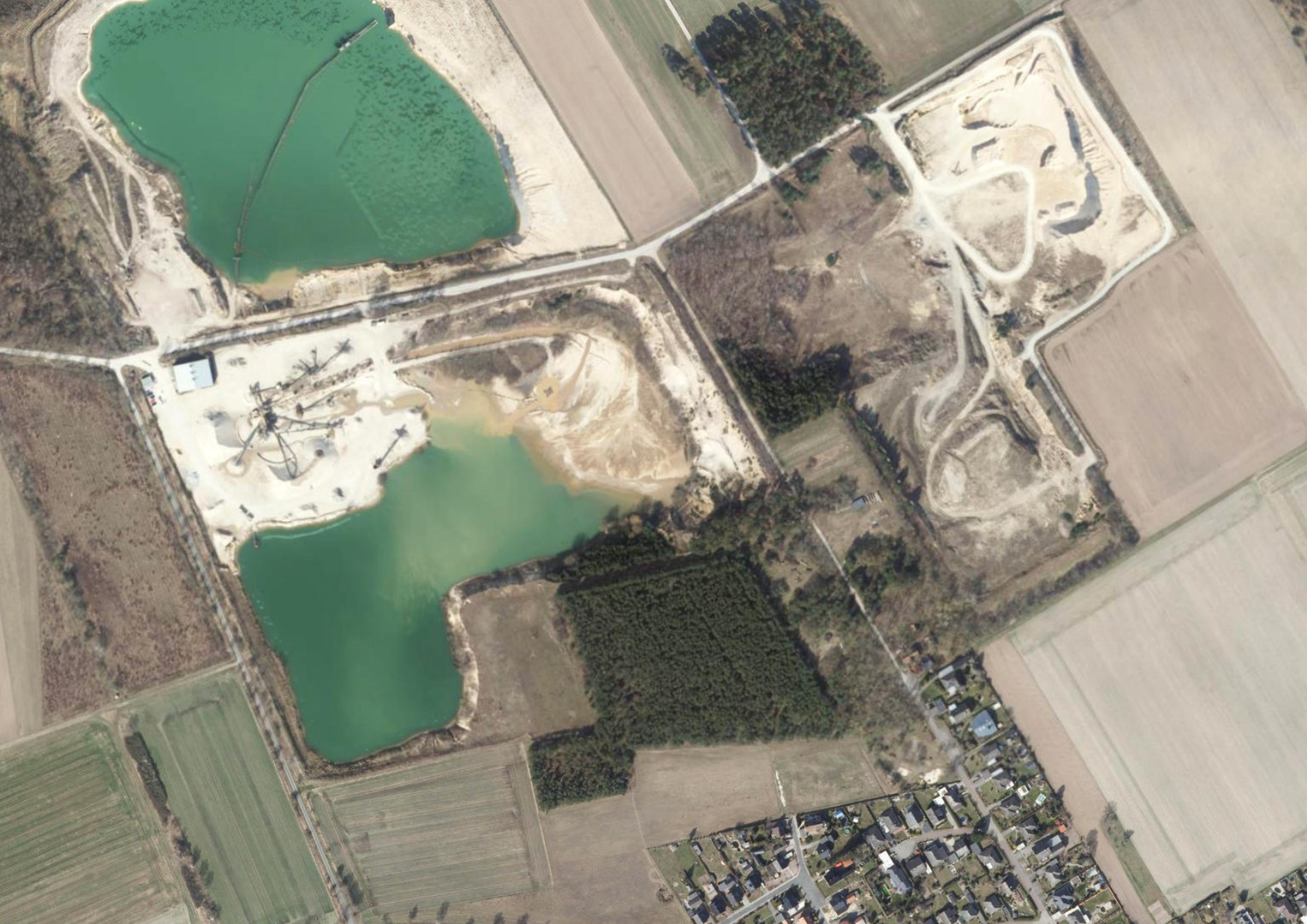}

Kieswerk Bahrdorf is a \textcolor{clueOrange}{producer and wholesaler} of bulk materials such as
\textcolor{clueGreen}{sand and gravel}, supplying the greater Wolfsburg area.
\end{minipage}
&
\begin{minipage}[t]{\linewidth}
\includegraphics[height=\linewidth, angle=90]{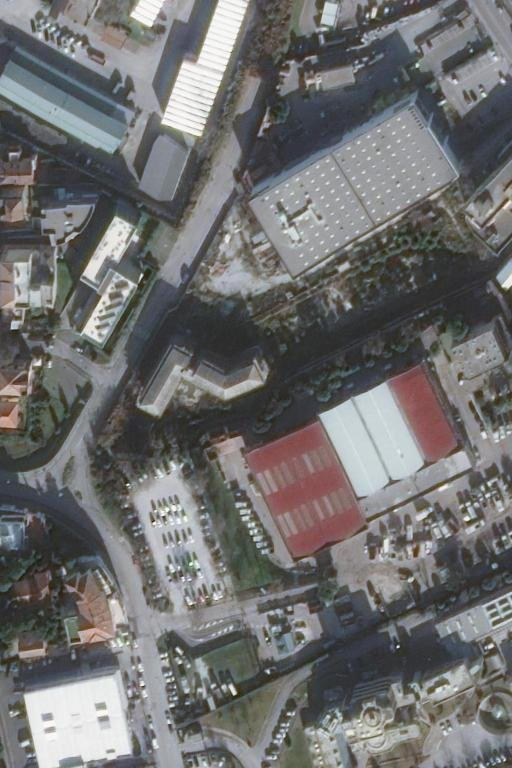}

AnconAmbiente provides \textcolor{clueGreen}{urban waste collection and environmental services}
for municipalities in the Province of Ancona.
\end{minipage}
\\[6pt]

\textbf{Clues} & \textbf{Clues} \\

\begin{minipage}[t]{\linewidth}
\textbf{Satellite:}
\textcolor{clueGreen}{[quarrying]} Excavation and heavy machinery. \\
\textbf{Website:}
\textcolor{clueOrange}{[wholesale]} Producer and wholesaler of bulk materials.
\end{minipage}
&
\begin{minipage}[t]{\linewidth}
\textbf{Satellite:}
\textcolor{clueOrange}{[manufacturing]} Large industrial buildings. \\
\textbf{Website:}
\textcolor{clueGreen}{[waste]} Waste collection services.
\end{minipage}
\\[6pt]

\textbf{Rationale and Label} & \textbf{Rationale and Label} \\

\begin{minipage}[t]{\linewidth}
\textcolor{clueGreen}{Satellite imagery} shows excavation and heavy machinery consistent with
quarrying. The term ``Kieswerk'' explicitly refers to a gravel pit. \\
\textbf{Label:} Mining and Quarrying
\end{minipage}
&
\begin{minipage}[t]{\linewidth}
Observed infrastructure and \textcolor{clueGreen}{website content} indicate organized waste
collection rather than manufacturing. \\
\textbf{Label:} Water Supply; Sewerage, Waste Management and Remediation Activities
\end{minipage}
\\

\bottomrule
\end{tabular}

\caption{Qualitative MONETA examples for correct inferences with contradicting clues Satellite image and website summary are followed by extracted clues. LLM clues, rationale, and decision are obtained via InternVL 3-38B in the mixture configuration. \textcolor{clueGreen}{Green} indicates content supporting ground truth while \textcolor{clueOrange}{orange} contents do not match the ground truth label.}
\label{tab:qualitative_moneta_example}
\end{table}

\begin{table}[!h]
\scriptsize
\resizebox{\linewidth}{!}{%
\begin{tabular}{lrr}
\toprule
\textbf{Configuration} & \textbf{Company Websites} & \textbf{NAICS} \\
\midrule
\multicolumn{3}{l}{Zero-Shot} \\
--- 7B & 57.93 & 50.19 \\
--- 32B & 62.79 & 57.45 \\
 \midrule
\multicolumn{3}{l}{Few-Shot}\\
--- 7B & 58.16 & 51.22 \\
--- 32B  & 68.86 & 56.72 \\
\midrule
\multicolumn{3}{l}{OURS LORA - 7B}\\
--- Company Website & \textbf{89.74} & 15.62 \\
--- ExioNAICS & 15.76 & 61.44 \\
\midrule
\multicolumn{3}{l}{\citet{guo_group_2025}}\\
--- MiniLML3 &  \xmark & 89.73 \\
--- MpnetBase & \xmark & \textbf{91.73} \\
\midrule
\multicolumn{3}{l}{\citet{jagric_ai_2024}}\\
--- BERT  & 88.23 & \xmark\\
\bottomrule
\end{tabular}%
}
\caption{Qwen 2.5 accuracies on text-only benchmarks: ExioNAICS \cite{guo_group_2025} and Company Websites \cite{jagric_ai_2024}.}
\label{tab:ablation_text_only_benchmarks}
\end{table}

\noindent \textbf{Text Only Benchmarks:} We validated our methodology on publicly available text-only benchmarks ExioNAICS \cite{guo_group_2025} and Company Websites \cite{jagric_ai_2024}. For reproducibility, we followed their guidelines and used a fixed seed for data splitting, as shown in Appendix \ref{sec:implementation}. As they fine-tuned models, we included test results for few shots (1 sample per class) and adapted models (with LORA) in Table~\ref{tab:ablation_text_only_benchmarks}. We used Qwen 2.5 text model as it is text core for both QwenVL 2.5 and InternVL 3. 

With zero-shot pipeline, we observed similar performances for both datasets. We had minimal gains with few-shot for the company websites dataset and no improvement for NAICS-2. After fine-tuning our models with LORA, we surpassed \citet{jagric_ai_2024} on their task. 

Fine-tuning models to a fixed classification scheme makes models fragile to future revisions. To demonstrate this, we evaluated adapted models on an alternative task and observed a performance drop exceeding \textbf{35\%} relative to zero-shot inference. Not only does fine-tuning require a significant amount of labeled data, but fine-tuned models are also not usable after classification schemes change. As noted by \citet{guo_group_2025}, industry classifications have changed in form multiple times over the years. In contrast, our prompting strategy remains adaptable and robust to such updates.

\section{Conclusion}
In this work, we introduce \texttt{MONETA}, a new dataset and task for multimodal industry classification. Our benchmark reflects real-world challenges in business registers by enabling industry classification using satellite and OSM imagery with external resources and NACE labels.

We proposed a multi-turn pipeline that generates clues from each resource and introduces metrics for their quantitative analysis. We validated our pipeline on two existing unimodal datasets and outperformed one configuration. Our experiments highlighted the limitations of fine-tuned models in cross-domain settings and the robustness of our zero-shot alternative. \texttt{MONETA} reveals the difficulty of the task and the textual bias of MLLMs.

Beyond our methodology, \texttt{MONETA} is relevant to policymakers and financial experts by supporting financial risk assessment, market analysis, and regional economic monitoring. We enable fast and reliable industry identification for newly founded or data-sparse entries in business registers. Future work will analyze its integration into real-world decision-making.

\section*{Limitations}
During this study, we use Gemini and ChatGPT to automatically create mappings from NACE to OSM tags. This process may introduce errors in the dataset preparation stage. In order to increase data quality, we have done extensive manual evaluation referring to the OSM wiki and NACE official guidelines. 

During the data preparation of this work, NACE received another revision named as NACE Rev.~2.1. This revision split one of the major categories. Unlike prior fine-tuning approaches, our prompts can be easily modified to the new scheme and tested accordingly. 


One of the limitations regarding the MLLM experiments is that some of the entities, due to initial filtering for external resources, may be in the training corpora as we include Wikidata and Wikipedia. This may be the reason behind the initially high accuracies.

We believe that future research can benefit from expert feedback and annotation in initial mapping and data quality assurance. Furthermore, the current setup can be tested with adapted MLLMs for financial and geospatial domains.

\section*{Ethics Statement}

\noindent \textbf{Social Impact: }This work provides multimodal benchmark for industry classification task. Company entries are selected from OpenStreetMap which is publicly available. \texttt{MONETA} is intended solely for research purposes. 

\noindent \textbf{Dataset Access: }Our code and dataset annotations are released under the Apache 2.0 and CC BY-SA 4.0 licenses, respectively. We do not hold the rights for ESRI ArcGIS World Imagery and thus will not distribute satellite images obtained from the tiles. In our datasets, we will release OSM tags and images licensed Open Data Commons Open Database License (ODbL) which also contain the bounding boxes and links to the external sources (Wikidata, Wikipedia and websites). We will also release the script to retrieve tiles and external content.

\noindent \textbf{AI Assistants: }AI assistants are used in this work to assist with writing by correcting grammar and code by prompt optimization and debugging.

\section*{Acknowledgements}
This work has been supported by the Research Center Trustworthy Data Science and Security \url{(https://rc-trust.ai)}, one of the Research Alliance centers within the \url{https://uaruhr.de}. 

\bibliography{bib}
\bibliographystyle{acl_natbib}

\appendix
\section{Dataset Properties}\label{appendix:data_properties}
\texttt{MONETA} contains 1,000 businesses in Europe with EU Guidelines' NACE economic activity labels. Each entry contains two geospatial and at least one textual resource. As the NACE section T, Activities of Households as Employers; Undifferentiated Goods- and Services-producing Activities of Households for Own Use, cannot be obtained through OSM, we used the remaining 20 NACE sections for economic activities. For each section, we list 50 entities. 

In Table~\ref{tab:dataset_attributes}, we listed data fields in our dataset. After filtering OSM, we identified NACE code and added category field. We extracted id, name, type and bbox from OSM fields. All the remaining attributes of OSM are present in OSM tags. After retrieving the images, we included image paths for reproducibility. We also obtained text resources (website, Wikidata or Website) and added as additional field. 

\begin{figure*}[!h]
    \centering
    \includegraphics[width=\linewidth]{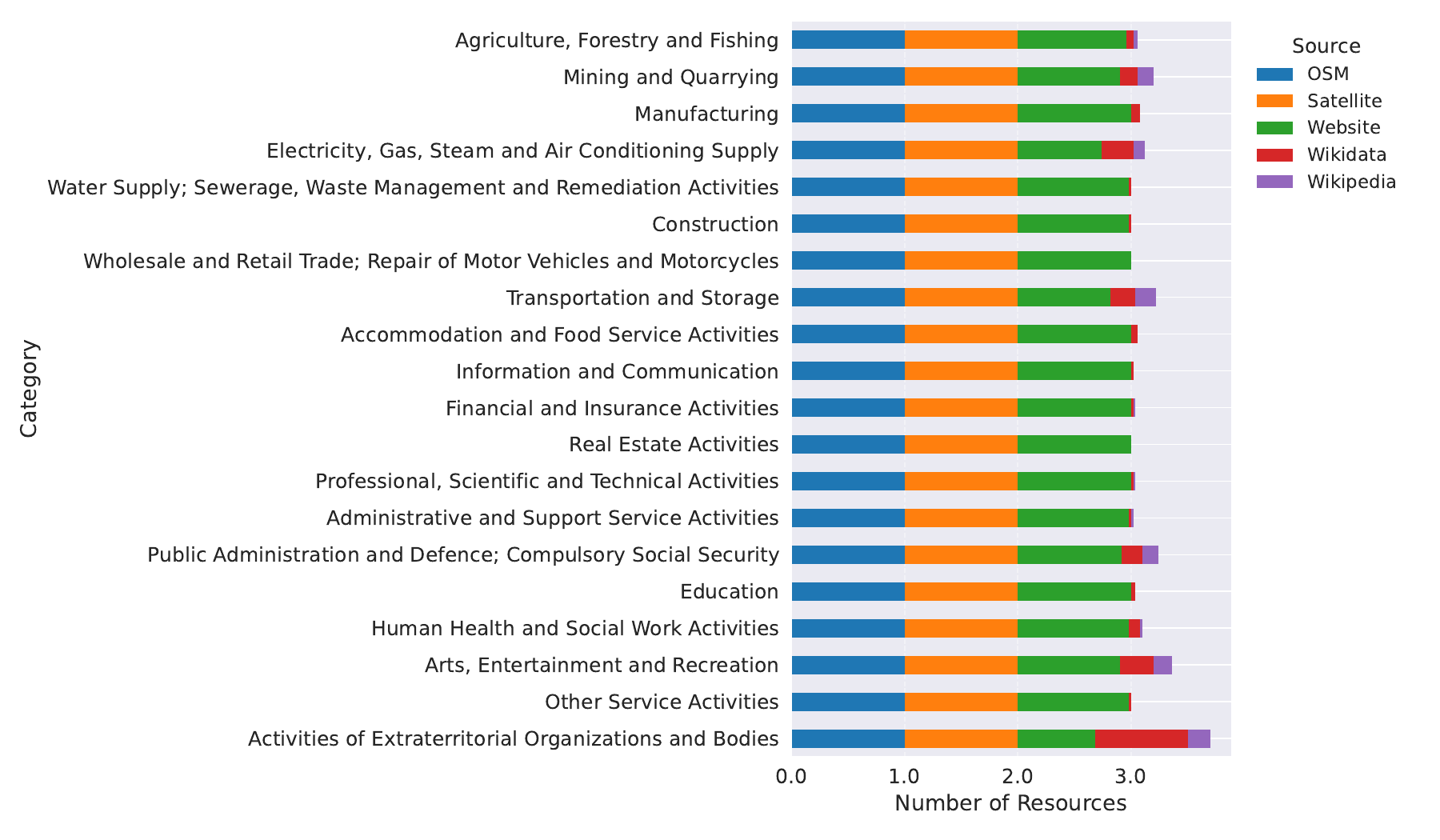}
    \caption{Average number of associated resources (OSM images, satellite images, website text, Wikidata, and Wikipedia) per NACE sections in the \texttt{MONETA} dataset }
    \label{fig:dataset_resources_per_category}
\end{figure*}

\begin{table}[h!]
\centering
\footnotesize
\begin{tabularx}{\columnwidth}{|l|X|}
\hline
\textbf{Attribute} & \textbf{Description} \\
\hline
\texttt{id} & Unique identifier for the object. \\
\texttt{type} & The object type (e.g., node, way, relation). \\
\texttt{name} & Human-readable name given in OSM. \\
\texttt{bbox} & Bounding box representing the spatial extent of the object. \\
\texttt{osm\_tags} & OpenStreetMap tags associated with the object. \\
\texttt{category} & NACE Rev 2 sector classification of the entity (A to U). \\
\texttt{image\_paths} & Dictionary of image paths for satellite and OSM images. \\
\texttt{sources} & Dictionary of external sources (Website Text, Wikipedia Text, Wikidata JSON). \\
\hline
\end{tabularx}
\caption{Dataset entry contents. Each entry of \texttt{MONETA} contains these attributes. Id, type, name, bbox and OSM tags are retrieved from OSM. Sources are retrieved online from existing OSM tags.}
\label{tab:dataset_attributes}
\end{table}

\begin{table}[ht]
\centering
\small
\renewcommand{\arraystretch}{1.2}
\begin{tabular}{|p{2cm}|p{5cm}|}
\hline
\textbf{Attribute} & \textbf{Value} \\
\hline
id & 122563530 \\
\hline
type & way \\
\hline
name & Heim Kieswerk \\
\hline
bbox &
[12.4893727, 50.9761359, 12.5089029, 50.9916218] \\
\hline
category & B (Mining and quarrying) \\
\hline
osm\_tags &
addr:city = Nobitz \newline
addr:country = DE \newline
addr:housenumber = 14c \newline
addr:postcode = 04603 \newline
addr:street = Altenburger Straße \newline
landuse = quarry \newline
resource = sand \newline
operator = Heim Kieswerk Nobitz GmbH \& Co. KG 
\\
\hline
image\_paths &
OSM: OSM\_PATH.png \newline
Satellite: Satellite\_PATH.png
\\
\hline
sources &
Website text extracted from \url{https://www.heim-gruppe.de} \newline
Wikipedia: -- \newline
Wikidata: --
\\
\hline
\end{tabular}
\caption{\texttt{MONETA} entry for \textit{Heim Kieswerk} derived from OpenStreetMap and external sources.}
\label{tab:heim_kieswerk}
\end{table}

A sample entry of \texttt{MONETA} used in qualitative analysis has the attributes given in Table~\ref{tab:heim_kieswerk}. This entry contains only the website as the external source. Due to the content length, we provided the link instead. We replaced image paths with placeholders.  
\begin{figure*}[!h]
    \centering
    \includegraphics[width=\textwidth]{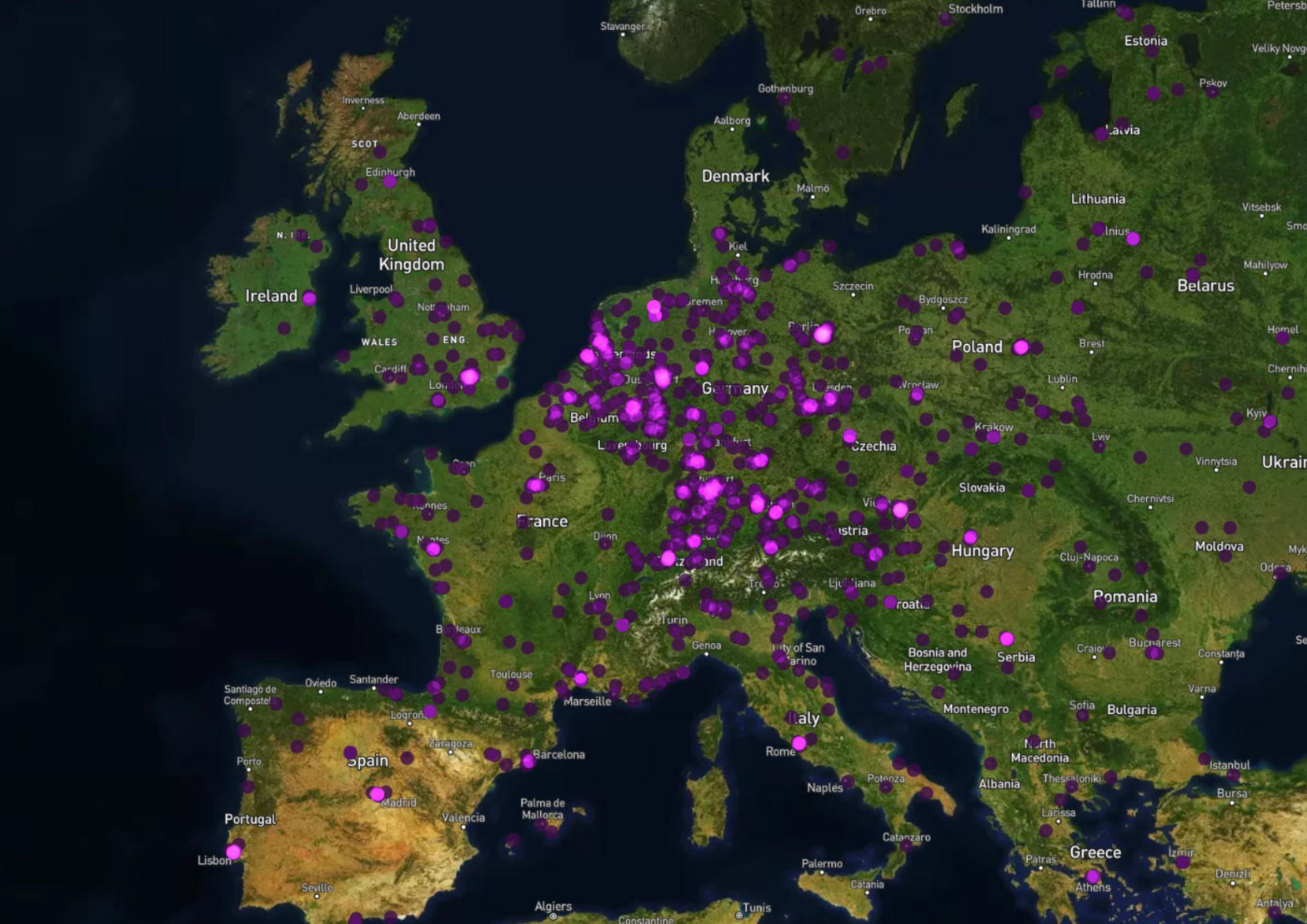}
    \caption{Spatial distribution of \texttt{MONETA} entries across Europe, showing the geographic coverage of OSM-derived entities used for NACE classification.}
    \label{fig:moneta_europe}
\end{figure*}

\subsection{NACE to OSM Mapping}\label{sec:appendix_nace_osm}

OSM contains tags describing the geospatial entity. These tags can indicate contact information, address, external database references (such as Wikidata), building structures, and many economic properties. However, there is no one-to-one mapping connecting OSM to any existing industry classification framework. As far as we are concerned, this study is the first connecting OSM tags to the NACE industry classification scheme. 

In this section, we illustrate the data preparation workflow shown in Figure~\ref{fig:dataset_construction} using NACE Rev.2 Section K as an example. We first extract the official NACE guideline from the RDF/XML source. Extracted fields are title, content, scope, additional content, and exclusions as shown in Figure~\ref{fig:appendix_nace_xml_k}.

These textual descriptions are then provided to Gemini to generate a candidate list of OpenStreetMap (OSM) tags relevant to the economic activities defined under Section K (Figure~\ref{fig:appendix_gemini_k}). As Gemini can hallucinate the tags or recommend rarer tags in the OSM database, we do not use the generations directly. Instead, we verify the generated tags list, one by one, to ensure they exist and fit the scope of the related economic activity. We discard the tags that are non-standard, weakly related, or rarely used. We also add relevant tags from the OSM TagInfo database matching the NACE description. 

For example, \textcolor{orange}{company=insurance} is a rare tag with 27 entries around the globe. \textcolor{orange}{office=company}, on the other hand, is a broad tag that can correspond to any company that may not have an economic activity related to Financial and Insurance activity. \textcolor{darkgreen}{office=financial} is a valid and common tag fitting to section K. Furthermore, it is often used with address tags, which helps the data quality assurance. 

\begin{figure*}[!h]
    \centering
    \includegraphics[width=\linewidth]{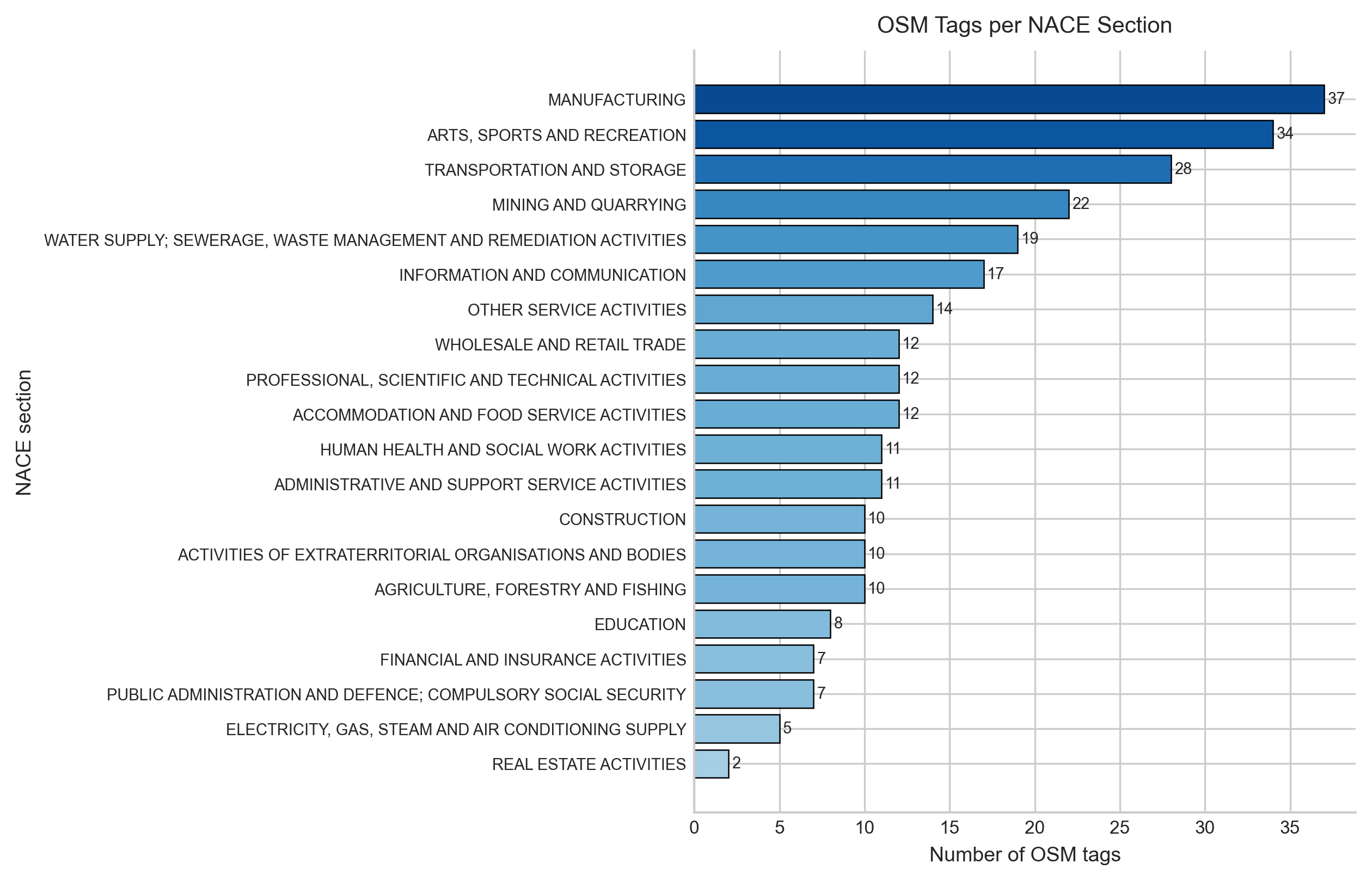}
    \caption{Distribution of OpenStreetMap (OSM) tags across NACE economic activity sections. The horizontal bar chart illustrates the mapping density for each category, with bar length and labels indicating the total number of unique OSM tags associated with a specific NACE section.}
    \label{fig:osm_to_nace}
\end{figure*}

After a thorough qualitative assessment, we obtained a list of OSM tags for each NACE section. The number of OSM tags per section is shown in Figure~\ref{fig:osm_to_nace}. The highest numbers of tags are in sections Manufacturing (C), Arts, Sport and Recreation (R), and Transportation and Storage (H). However, the number of unique OSM tags does not translate into the number of elements, as each tag has a different number of entities in OpenStreetMap. For example, \textcolor{darkgreen}{office=estate\_agent}, belongs to real estate activities, contains 97,716 objects in OSM, whereas \textcolor{darkgreen}{industrial=textile} of manufacturing contains 200 objects. With our data, we also release the mapping scheme for further research.

\subsection{MONETA versioning}
After creating the NACE to OSM Mapping, we retrieve the elements from OSM's most recent European extract. To ensure data quality, we limit the search to entries with a name tag in OSM. We iteratively add filters to generate versions of our datasets:
\begin{itemize}
    \item We first use NACE-induced tags and the name tag to generate the bronze version. In this version, not all entries carry address information and external resource pointers.
    \item We, then, use address tags to filter the bronze version and form the silver dataset. 
    \item Finally, we used Wikidata, Wikipedia, and Website tags to ensure at least one external resource would be available. This version is named the gold version. 
\end{itemize}

From the gold version, we sampled two versions for experiments:
\begin{itemize}
    \item \texttt{MONETA}: A uniform dataset with 50 entries per category. Each entry has an address, two images from GIS, and at least one external text resource. Overall, this benchmark contains 1,000 businesses in Europe.
    \item \texttt{MONETA-10K}: Extended version of the original dataset. It contains 10,000 businesses with the same multimodal attributes. This version is not uniform in terms of NACE categories.
\end{itemize}

\begin{figure*}[!h]
\centering
\begin{tcolorbox}[
  colback=gray!5,
  colframe=gray!70!black,
  title=\textbf{NACE Rev.2 RDF/XML Extract — Section K},
  fonttitle=\bfseries,
  sharp corners,
  boxrule=0.8pt,
  width=0.95\textwidth
]

\textbf{Source}\\
Retrieved from:\\
\url{https://publications.europa.eu/resource/authority/ux2/nace2/K}

\vspace{0.5em}
\textbf{Parsed Content}

\begin{tcolorbox}[
  colback=white,
  colframe=gray!50,
  boxrule=0.5pt,
  sharp corners
]
\ttfamily\footnotesize
\begin{verbatim}
{
  'Official Name': 'K FINANCIAL AND INSURANCE ACTIVITIES',
  'Alternative Name': 'FINANCIAL AND INSURANCE ACTIVITIES',
  'Scope': None,
  'Content': 'This section includes financial service activities,
  including insurance, reinsurance and pension funding
  activities and activities to support financial services.',
  'Additional Content': 'This section also includes the activities
  of holding assets, such as holding companies and trusts,
  funds and similar financial entities.',
  'Exclusion': None
}
\end{verbatim}
\end{tcolorbox}

\end{tcolorbox}
\caption{RDF/XML NACE official guideline content for Section K (Financial and Insurance Activities).}
\label{fig:appendix_nace_xml_k}
\end{figure*}

\begin{figure}[!h]
\centering
\begin{tcolorbox}[
  colback=green!3,
  colframe=green!50!black,
  title=\textbf{Gemini-Generated OSM Tags \\ Section K},
  fonttitle=\bfseries,
  sharp corners,
  boxrule=0.8pt,
  width=0.95\columnwidth
]
\begin{itemize}
  \item amenity=bank
  \item amenity=atm
  \item \textbf{\textcolor{darkgreen}{amenity=bureau\_de\_change}}
  \item shop=money\_lender
  \item shop=insurance
  \item \textbf{\textcolor{darkgreen}{office=financial}}
  \item \textbf{\textcolor{darkgreen}{office=insurance}}
  \item office=financial\_advisor
  \item office=company
  \item company=insurance
  \item office=consulting
\end{itemize}

\end{tcolorbox}
\caption{Gemini-generated OpenStreetMap tag candidates for the NACE Financial and Insurance Activities section (K). \textcolor{darkgreen}{Relevant tags} align with official NACE definitions, while other tags are either rare, non-standard, or weakly related.}
\label{fig:appendix_gemini_k}
\end{figure}

\subsection{NACE Details}
We extracted NACE codes, titles, descriptions and keywords for prompting. Summary of these attributes are given in Table~\ref{tab:appendix_nace_content}.

\begin{table*}[!h]
    \centering
    \scriptsize
    \begin{tabular}{|lp{3.5cm}|p{6.5cm}|p{3cm}| }
    \hline
    \toprule
        \textbf{Section} & \textbf{Title} & \textbf{Description} & \textbf{Keywords} \\ \midrule
        A & Agriculture, Forestry and Fishing & This section covers the utilization of plant and animal natural resources through farming, animal husbandry, and harvesting from natural environments. & [agriculture], [forestry], [fishing], [crops], [livestock], [timber] \\ 
        B & Mining and Quarrying & This section includes the extraction of naturally occurring minerals in solid, liquid, or gaseous forms, using various methods such as underground mining, surface mining, and well operations, along with related preparation activities. & [mining], [quarrying], [oil], [coal], [ores]\\ 
        C & Manufacturing & This section includes the physical or chemical transformation of raw materials or components into new products, typically resulting in outputs ready for use or as inputs to further manufacturing. & [manufacturing], [processing], [assembly], [fabrication]\\ 
        D & Electricity, Gas, Steam and Air Conditioning Supply & This section covers the provision and distribution of electricity, natural gas, steam, hot water, and air conditioning through a permanent infrastructure of networks such as lines, mains, and pipes. &  [electricity], [gas], [steam] \\ 
        E & Water Supply; Sewerage, Waste Management and Remediation Activities & This section includes the collection, treatment, and disposal of waste and sewage, as well as the management of contaminated sites and the supply of water for various uses. & [water], [sewerage], [waste], [remediation]\\ 
        F & Construction & This section covers general and specialised construction activities for buildings and civil engineering works, including new projects, repairs, additions, and temporary structures, whether performed directly or through subcontracting. & [construction], [building], [infrastructure]\\ 
        G & Wholesale and Retail Trade; Repair of Motor Vehicles and Motorcycles & This section includes the wholesale and retail sale of goods without transformation and related services, as well as the repair of motor vehicles and motorcycles. & [wholesale], [retail], [trade], [resale], [vehicle-repair]\\ 
        H & Transportation and Storage & This section includes the transport of passengers or freight by various modes, along with related services such as cargo handling, storage, and postal and courier activities. & [transport], [logistics], [freight], [storage], [postal]\\ 
        I & Accommodation and Food Service Activities & This section covers short-term accommodation services for travelers and the preparation and serving of meals and drinks for immediate consumption. & [accommodation], [hotels], [restaurants], [catering]\\ 
        J & Information and Communication & This section includes the creation, publishing, and distribution of information and cultural content, telecommunications, IT services, and data processing activities. & [information], [communication], [telecom], [publishing], [IT]\\ 
        K & Financial and Insurance Activities & This section includes activities related to financial services, insurance and pension funding, and asset-holding entities such as holding companies and trusts. & [finance], [insurance], [banking], [investment]\\ 
        L & Real Estate Activities & This section includes activities related to real estate sales, rentals, management, and related services, carried out either on owned or leased property or on a contract basis. & [real-estate], [property], [leasing]\\
        M & Professional, Scientific and Technical Activities & This section includes specialised services requiring high levels of expertise, such as legal, accounting, engineering, and scientific research services. & [professional], [scientific], [technical], [legal], [engineering], [research]\\
        N & Administrative and Support Service Activities & This section includes support services for general business operations that do not primarily involve the transfer of specialised knowledge, such as employment services, security, and facility management. & [administration], [support], [employment], [security], [cleaning]\\ 
        O & Public Administration and Defence; Compulsory Social Security & This section includes government-related activities such as legislation, taxation, national defence, public order, immigration, foreign affairs, and compulsory social security administration. & [government], [defence], [legislation], [taxation]\\ 
        P & Education & This section includes all levels and types of education, from preschool to higher education, including adult and special education, whether provided publicly or privately, through various formats such as in-person or online. & [education], [training], [schooling]\\ 
        Q & Human Health and Social Work Activities & This section includes medical care by health professionals, residential care involving health support, and social work activities without health care involvement. & [health], [social-care], [medical], [hospitals], [clinics] \\ 
        R & Arts, Entertainment and Recreation & This section includes cultural, artistic, entertainment, and recreational activities for the general public, including live shows, museums, gambling, sports, and leisure facilities. & [arts], [entertainment], [recreation], [sports], [culture] \\
        S & Other Service Activities & This section includes a variety of personal services not classified elsewhere, such as those provided by membership organisations and the repair of computers and household goods. & [personal-services], [household-services], [memberships], [repairs] \\
        T & Activities of Households as Employers; Undifferentiated Goods- and Services-producing Activities of Households for Own Use & This section includes households’ subsistence production of goods and services for their own use, when no primary activity can be identified and the output is not for market sale. & [household-employment], [household-production]\\ 
        U & Activities of Extraterritorial Organisations and Bodies & This section includes the activities of international organisations such as the UN, IMF, World Bank, and diplomatic missions determined by the host country location. & [extraterritorial], [embassies], [diplomacy]\\ \bottomrule
    \end{tabular}
    \caption{NACE Section Codes, Titles, AI-generated descriptions and keywords (from official guidelines). During the multi-turn inference, MLLMs will generate economic activity clues based on provided keywords.}
    \label{tab:appendix_nace_content}
\end{table*}

\section{MONETA-10K}\label{appendix:moneta10k}
Our NACE to OSM mapping allows us to retrieve elements for more than the 1,000 businesses we used in this study. However, due to computational resources and budgeting, we can test various input configurations and experiment dimensions with the released version of \texttt{MONETA}. Therefore, we sampled 50 entries per NACE section. However, using the same mapping, we also generated a more comprehensive benchmark which we call \texttt{MONETA-10K}. This benchmark, as the name implies, contains 10,000 businesses with NACE section labels. 

\subsection{Dataset Details}

\begin{figure}[!h]
    \centering
    \includegraphics[width=\linewidth]{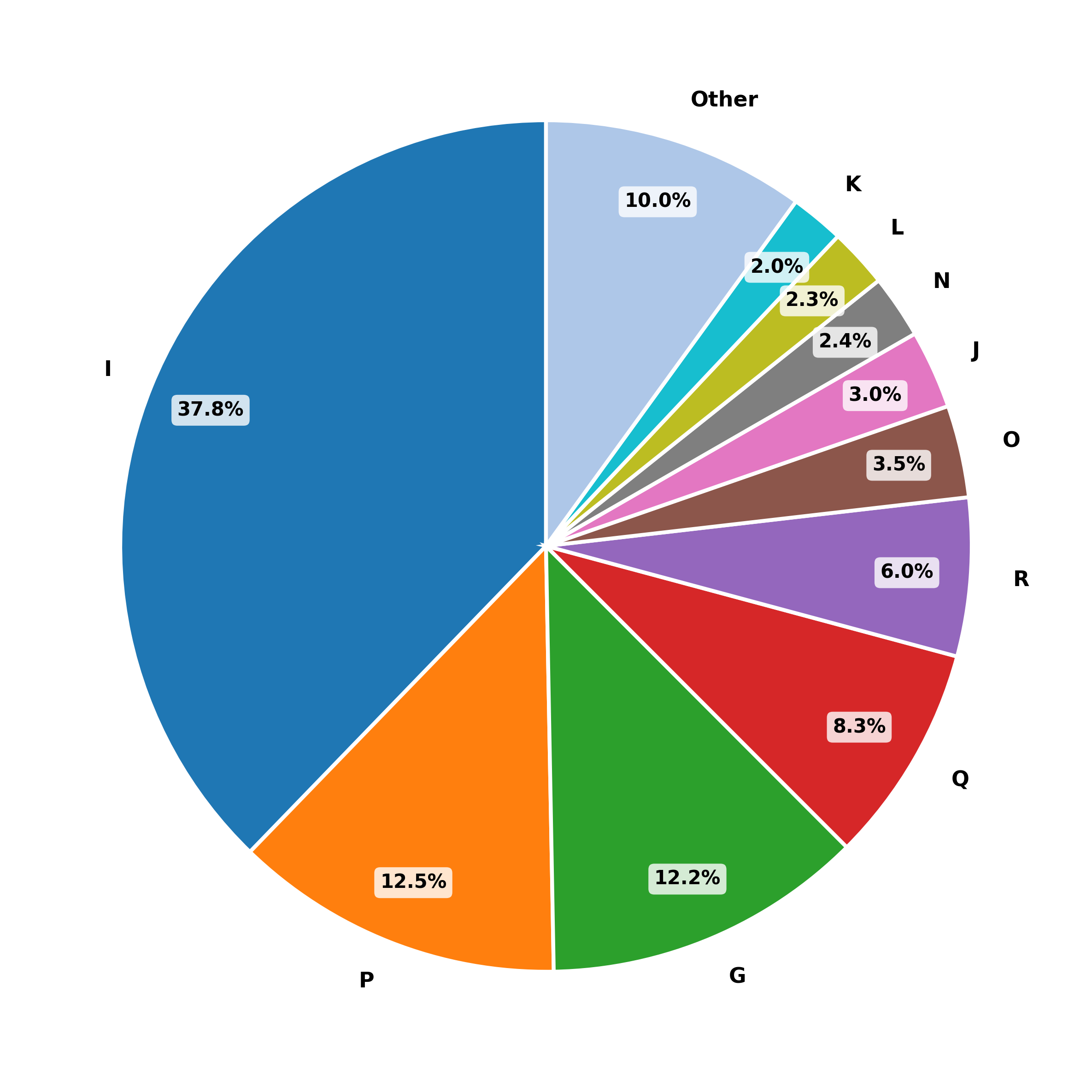}
    \caption{NACE section distribution for \texttt{MONETA-10K}. Sections less than 2\% are grouped into other for demonstration.}
    \label{fig:moneta10k_pie}
\end{figure}

In this version, all entries possess at least 1 external resource and 2 geospatial images. The distribution of sections is given in Figure~\ref{fig:moneta10k_pie}. Furthermore, we provided details of external resources in Table~\ref{tab:moneta10k_external}.

\begin{table}[!h]
\centering
\begin{tabular}{l|c}
\toprule
Text Source & Entry Count  \\ \midrule
Website       &                   9,015 \\
Wikidata      &                       276\\
Wikipedia       &                      13\\
Wikidata + Website   &                315\\
Wikidata + Wikipedia    &             147\\
Wikipedia + Website    &               13\\
All &     221\\\bottomrule
\end{tabular}
\caption{\texttt{MONETA-10K} external resource counts. All denotes the existence of Wikidata, Wikipedia, and Website}
\label{tab:moneta10k_external}
\end{table}

\subsection{Comparison with \texttt{MONETA} Results}
We examined MLLM performance on \texttt{MONETA-1OK} using the InternVL 3 - 8B model in our baseline configuration. We used text output with single token inference and provided only the NACE sections and titles in the system prompt. We used all available resources per entry. 

In Table~\ref{tab:moneta-10k-comparison}, we demonstrated results for macro and weighted F1-Score and Precision, Recall. The results for \texttt{MONETA} and \texttt{MONETA-10K} differ less than 5\% which indicates that \texttt{MONETA} can be used in NACE-based industry classification like a larger counterpart. 

\begin{table}[]
\centering
\begin{tabular}{lrr}
\toprule
Metric & \texttt{MONETA-10K} & \texttt{MONETA} \\
\midrule
Macro F1-score & 39.40 & 38.70 \\
Weighted F1-score & 45.90 & 40.70 \\
Precision & 52.30 & 52.30 \\
Recall & 39.40 & 39.70 \\
\bottomrule
\end{tabular}
\caption{Macro and Weighted F1, recall, and F1-score for the \texttt{MONETA-10K} and \texttt{MONETA} datasets with InternVL 8B using Zero-Shot pipeline, Text output, Simple prompt and all available resources.}
\label{tab:moneta-10k-comparison}
\end{table}

\section{Implementation Details}\label{sec:implementation}

\subsection{Frameworks}
In order to run models, we preferred Hugginface's Transformers, \cite{wolf-etal-2020-transformers} library due to multimodal support. During the ablation studies on text only benchmarks, we used Unlsoth \cite{unsloth} to train models with LoRA \cite{hu2022lora}.

\subsection{Infrastructure}
In our experiments, we used NVIDIA A100 40GB GPUs and increased number of GPUs depending on the model size. 

\subsection{Hyperparameters}
During the ablation studies for the text only benchmarks, we fine tuned models with LoRA \cite{hu2022lora} using rank 32 and alpha 64 for 5 epochs with learning rate $2e-4$.

\section{Additional Results}\label{sec:appendix}

\subsection{Section-wise Analysis}

\begin{figure*}[]
    \centering
    \includegraphics[width=\linewidth]{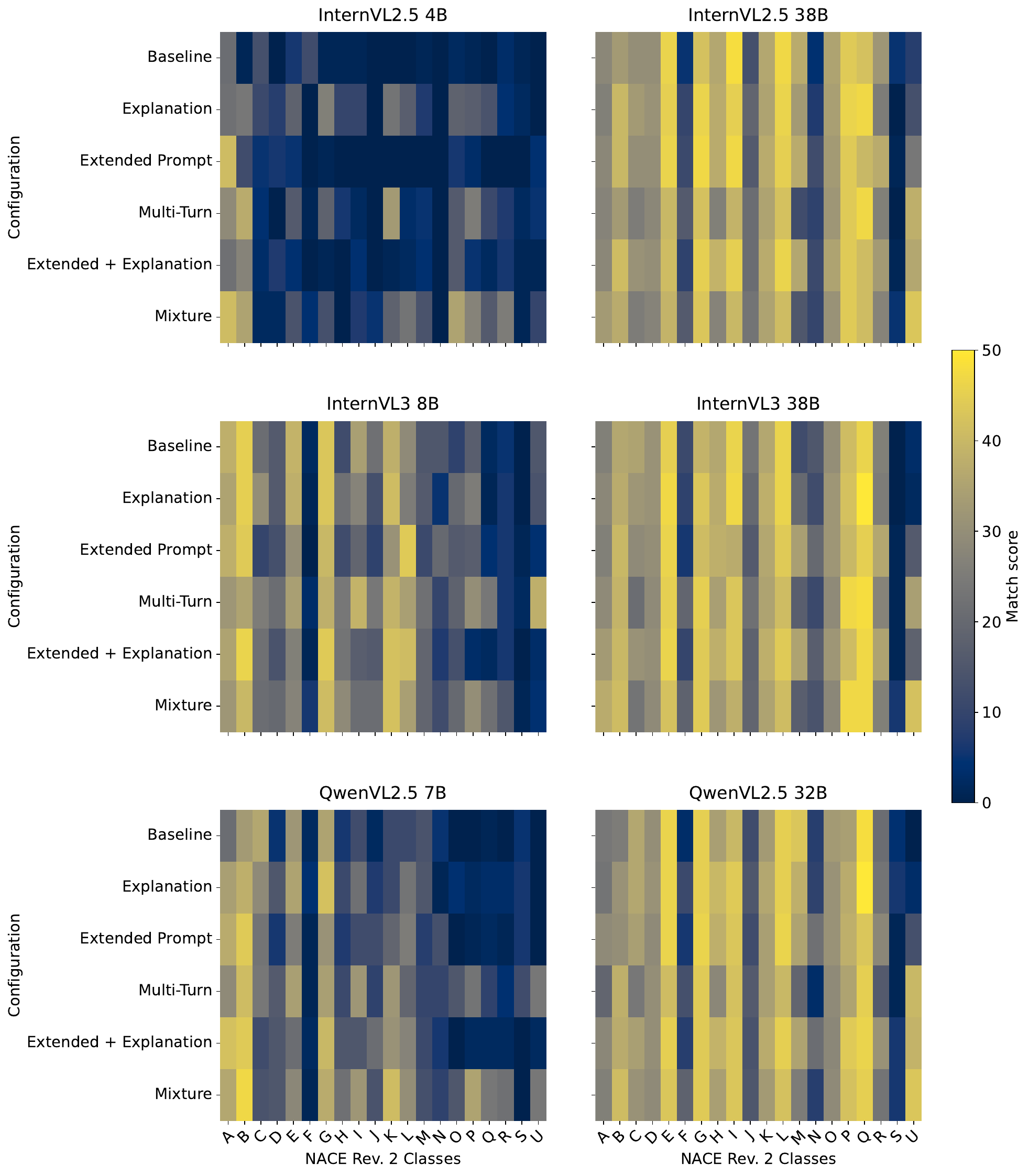}
    \caption{NACE Section-wise analysis for InternVL 2.5 (4B and 38B), InternVL 3 (8B and 38B), and QwenVL 2.5 (7B and 32B). Rows indicate experiment configurations. Mixture denotes Multi-Turn pipeline with classification explanations and prompt enrichment. Columns are the NACE section letters given in Table~\ref{tab:appendix_nace_content}.}
    \label{fig:appendix_confusion_matrix_config}
\end{figure*}

Based on the available configurations (Baseline, Explanation, Extended Prompt, Multi-Turn, Extended Prompt + (Explanation), Mixture), we created confusion matrices between ground truth and prediction results. We retrieved the counts from the diagonals and visualized them with respect to configurations in Figure~\ref{fig:appendix_confusion_matrix_config}. The smaller models performed poorly for the sections M (Professional, Scientific, and
Technical Activities) and S (Other Service Activities). The usage of multi-turn allowed smaller models to detect B (Mining and Quarrying) and U (Activities of Extraterritorial Organisations and Bodies). Larger models are overall consistent with the exceptions of sections F (Construction), N ( Administrative and Support Service Activities), and S (Other Service Activities). The effect of prompt context enrichment and multi-turn is also visible for U (Activities of Extraterritorial Organisations and Bodies) for larger models.

\subsection{Experiment Dimension Analysis}
\begin{figure*}[]
    \centering
    \includegraphics[width=\linewidth]{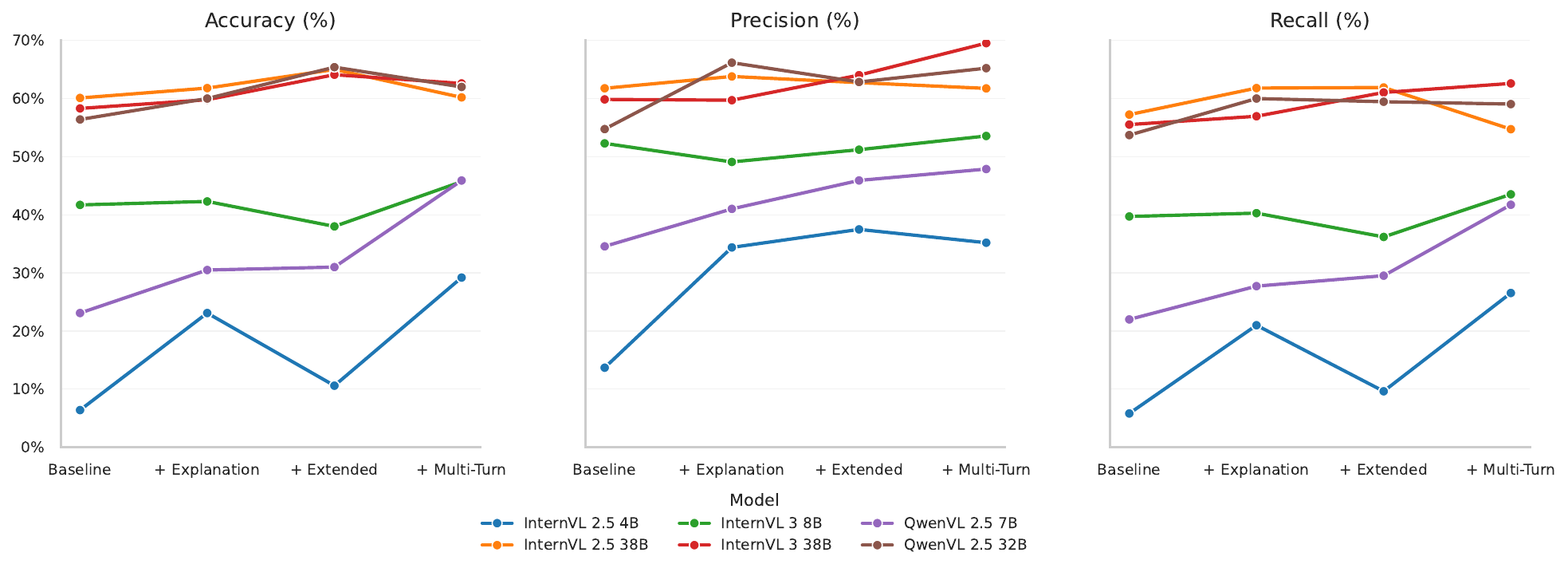}
    \caption{Experiment configuration results given in Accuracy, Precision, and Recall for InternVL 2.5 (4B and 38B), InternVL 3 (8B and 38B), and QwenVL 2.5 (7B and 32B). At each setting, the given configuration is added. The order of configurations is: Baseline, Explanation, Explanation + Extended Prompt, Mixture (with Multi-Turn).}
    \label{fig:combined_configurations}
\end{figure*}

We visualized the effect of experiment ensembles in our Figure~\ref{fig:combined_configurations}. In the x-axis, we incrementally added the results. Therefore, it corresponds to Baseline, Explanation, Extended Prompt + Explanation, and Multi-Turn + Extended Prompt + Explanation. In addition to accuracy, we used precision and recall. In all these metrics, changes in smaller models are superior compared to changes in larger models in the same family. 

\subsection{Clue Ablations}

\noindent \textbf{Model section preferences} Using the keywords in the freeform text, we grouped clue contents into NACE sections using the keyword list. Resulting groupings formed the clue frequency vectors. From clue frequency vectors, we identified percentages for each NACE sections for InternVL 3 (8 and 38B) and QwenVL 2.5 (7B and 32B). Regardless of the architecture and model size, we observe a strong representation of \textit{Wholesale and Retail Trade}, \textit{Transportation and Storage} and \textit{Accommodation and Food Service Activities}. In the original dataset, the distribution of visual sources are uniform. However for clues, bias is observed for the listed categories. In addition to this, as it was shown in Figure~\ref{fig:dataset_resources_per_category}, Wikidata and Wikipedia were dominant sources in the last category. While, smaller models fail to utilize these sources, larger models clearly extract correct clues. We visualized this in Appendix as a confusion matrix, In Figure~\ref{fig:clue_keywords_counts}.

\begin{figure*}[]
    \centering
    \includegraphics[width=\linewidth]{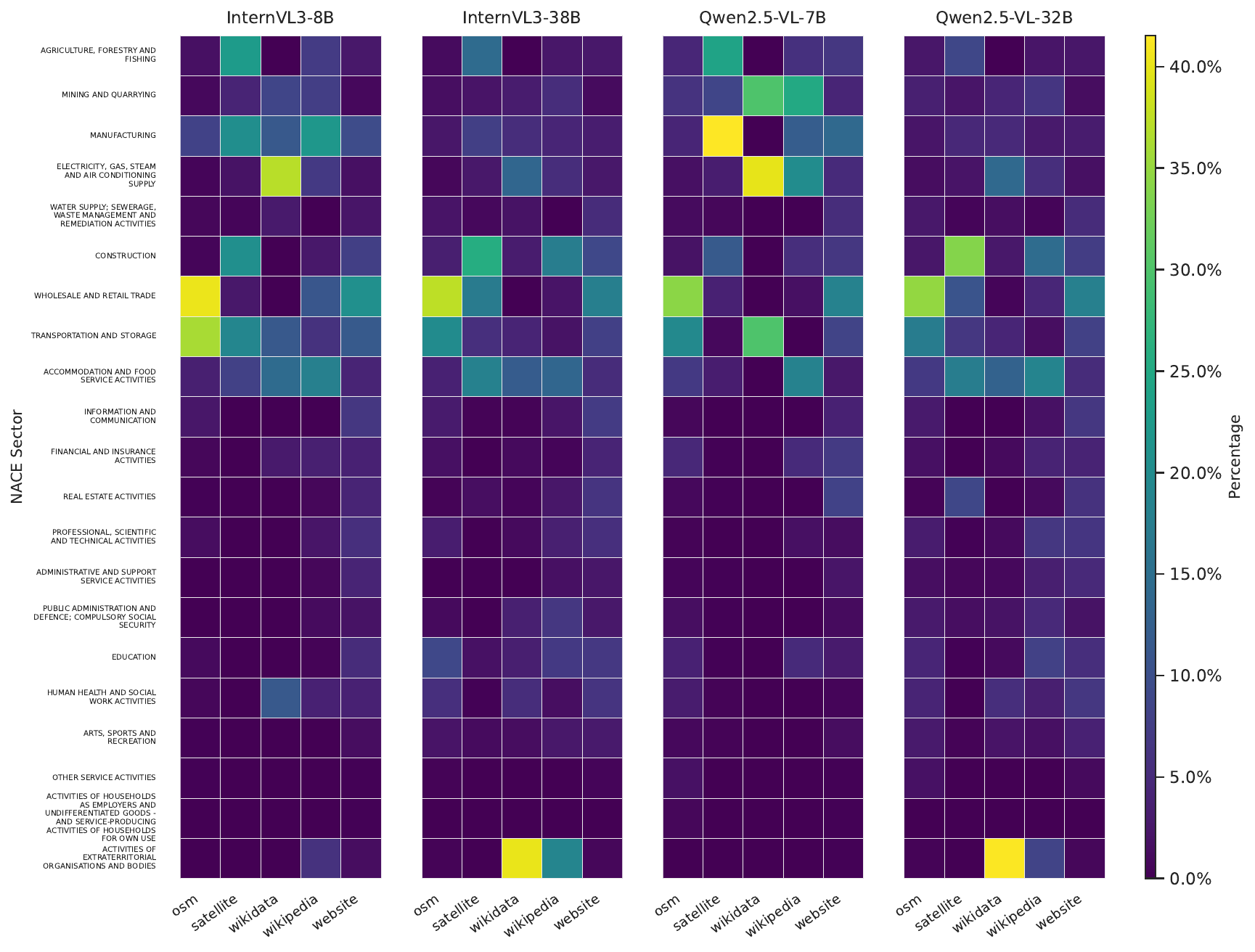}
    \caption{Clue Keywords confusion matrices obtained for InternVL 3 (8B and 38B), and QwenVL 2.5 (7B and 32B). Extracted keywords are grouped based on NACE sections in the rows. The columns are resources: OSM, Satellite, Wikidata, Wikipedia and Website.}
    \label{fig:clue_keywords_counts}
\end{figure*}
\begin{table*}[!h]
\centering
\resizebox{\textwidth}{!}{%
\begin{tabular}{llrrrrrrrrrr}
\toprule
\textbf{Model} &
  \textbf{Size (B)} &
  \multicolumn{2}{c}{\textbf{\textcolor{orange}{OSM}}} &
  \multicolumn{2}{c}{\textbf{\textcolor{orange}{Satellite}}} &
  \multicolumn{2}{c}{\textbf{Wikidata}} &
  \multicolumn{2}{c}{\textbf{Wikipedia}} &
  \multicolumn{2}{c}{\textbf{Website}} \\
  \midrule
\multirow{2}{*}{InternVL 2.5} & 4 & 1.74 & 2.62 & 6.26 & \textbf{8.46} & 6.56 & 0.82 & \textbf{14.54} & 7.85 & 7.61 & 6.47 \\
 & 38 & 8.15 & 16.05 & 6.58 & 8.12 & \textbf{54.10} & \textbf{56.28} & 37.58 & 38.83 & 30.03 & 43.45 \\ \midrule
\multirow{4}{*}{InternVL3} & 8 & 9.31 & 20.02 & 11.02 & 13.77 & 18.85 & 14.75 & 17.15 & 17.47 & \textbf{29.32} & \textbf{39.03} \\
&  14 & 10.22 &	19.55 & 7.83 & 	9.02 &	\textbf{40.57} & \textbf{47.95} & 31.25 &	37.11 &	34.23 &	45.69 \\
 & 38 & 14.77 & 23.40 & 11.04 & 14.62 & \textbf{58.61} & \textbf{57.65} & 33.92 & 41.23 & 34.58 & 45.72 \\
 &  78 & 10.58 & 13.56 & 9.73 & 11.34 & \textbf{57.38} & \textbf{55.33} & 35.08 & 34.62 & 39.18 & 51.04\\ \midrule
\multirow{3}{*}{Qwen VL 2.5} & 7 & 12.81 & 21.49 & 10.30 & 15.85 & 5.74 & 3.28 & 19.23 & 15.38 & \textbf{19.98} & \textbf{25.10} \\
 & 32 & 18.13 & 26.68 & 10.36 & 13.15 & \textbf{54.37} & \textbf{57.65} & 22.34 & 24.05 & 32.40 & 42.84 \\
 &  72 & 16.28 & 22.32 & 7.22 & 9.95 & \textbf{50.82} & \textbf{50.00} & 22.57 & 23.21 & 27.40 & 35.10 \\ \bottomrule
\end{tabular}%
}
\caption{Performance comparison in the Mixture setting, reporting correctness (left) and effectiveness (right) for each model across input modalities. \textcolor{orange}{Image}-based inputs are highlighted. \textbf{Bold} indicates the best performance in correctness/effectiveness for the model and size pair.}
\label{tab:clue_correctness_effectiveness}
\end{table*}

\begin{table*}[!h]
\scriptsize
\resizebox{\linewidth}{!}{%
\begin{tabular}{ll|rrrrr}
\toprule
  \textbf{Model} & 
  \textbf{Size (B)} &
  \textbf{\textcolor{orange}{OSM}} &
  \textbf{\textcolor{orange}{Satellite}} &
  \textbf{Wikidata} &
  \textbf{Wikipedia} &
  \textbf{Website} \\
\midrule
\multirow{2}{*}{InternVL 2.5} & 4 & 24.50 & 35.20 & 13.11 & \textbf{51.92} & 22.85 \\
& 38 & 59.20 & 20.70 & 71.31 & \textbf{84.62} & 77.47 \\
\midrule
\multirow{4}{*}{InternVL 3} & 8  & 84.60  & 77.50 & 31.15 & 67.31 & \textbf{91.82} \\
& 14 & 67.90  & 44.10 & 73.77 &82.69 & \textbf{90.65} \\
& 38 & 87.10  & 78.40 & 78.69 & \textbf{90.38} & 79.38 \\
& 78 & 73.40  & 30.00 & 74.59 & 80.77 & \textbf{85.76} \\
\midrule
\multirow{3}{*}{QwenVL 2.5} & 7  & \textbf{64.70}  & 47.10 & 11.48 & 44.23 & 60.26 \\
& 32 & \textbf{89.90}  & 84.90 & 75.41 & 76.92 & 78.75 \\
&  72 & \textbf{73.20}  & 30.50 & 60.66 & 53.85 & 68.33 \\
\bottomrule
\end{tabular}%
}
\caption{Information Discovery of inputs. \textcolor{orange}{Image} inputs are highlighted. \textbf{Bold} denotes the highest information discovery from a model-size pair.}
\label{tab:clue_information_discovery}
\end{table*}

We also analyzed the obtained clues using our correctness and effectiveness measures in Table~\ref{tab:clue_correctness_effectiveness}. For these experiments, we used open-source MLLMs InternVL 2.5 (4B and 38B), InternVL 3 (8B, 14B, 38B, and 78B) and QwenVL 2.5 (7B, 32B, and 72B). For the larger models, we observed that the highest correctness and effectiveness are attained via Wikidata context. The smaller models can utilize website context the most and thus achieve their highest effectiveness and correctness scores when using these data. Among the visual clues, OSM images appear to be more useful compared to satellite images. The clue effectiveness and correctness indicate that even for the best resource, Wikidata, metric performances are below 60\%. Thus, industry classification cannot be solved using only a single source.

\subsection{Information Discovery}
We instructed clue extractors to return No Economic Activity Found in case there is no evidence in the source. Using the number of inferences with this phrase, $NEI_c$, we defined \textbf{Information Discovery Rate} as:

\begin{align}
    \text{Information Discovery } \ (ID_c) &= 1 - \frac{NEI_c}{I_c}
\end{align}
$I_c$ denotes the total number of inferences per clue, $c$. The metric demonstrates the evidence retrieved from a input type $c$. It is scaled between 0 and 1.

\noindent \textbf{Information discovery from clues} In Table~\ref{tab:clue_information_discovery}, Information Discovery Rates are given. Smaller models of InternVL 2.5 and QwenVL 2.5 may fail to extract information. The information discovery highlights architectural differences. InternVL 2.5 and 3 discovered more information from texts more while QwenVL 2.5 discovered most of the information from OSM. Among the text sources, we observed that Website and Wikidata to contain more information regarding economic activities. Especially, InternVL 3 models generated economic activity clues for more than 80\% of the examples. According to the results, InternVL 3 has a stronger vision core compared to its predecessor, as it is now able to utilize satellite and OSM images more than \textbf{30\%} at the same size.

\section{Prompts}\label{sec:prompts}

\subsection{Data Preparation Prompts}

We use Gemini to create OSM tags list from NACE RDF/XML descriptions. This results are manually checked to create OSM Tags lists for each NACE section.
\begin{figure}[!ht]
\centering
\begin{tcolorbox}[
  colback=blue!3,
  colframe=blue!70!black,
  title=\textbf{NACE–OSM Tag Mapping Prompt},
  fonttitle=\bfseries,
  sharp corners,
  boxrule=0.8pt,
  width=0.95\columnwidth,
  left=6pt,
  right=6pt,
  top=6pt,
  bottom=6pt
]

\textbf{Task Description}

You will be given a description of a \textbf{NACE code}, representing a business activity.  
Your task is to identify relevant \textbf{OpenStreetMap (OSM)} tags that can be used to classify businesses or locations corresponding to this activity.

\vspace{0.6em}
\textbf{NACE Code Description:}
\begin{quote}
\{RDF/XML Extract\}
\end{quote}

\vspace{0.6em}
\textbf{Response Format}

Your response must consist \textbf{only} of a Python list of OSM tags, where each element is a string in the form \texttt{key=value}.

\begin{quote}
\texttt{["landuse=retail", "shop=supermarket", "amenity=parking"]}
\end{quote}

\vspace{0.6em}
\textbf{Constraints}
\begin{itemize}
  \item Every tag must include an \texttt{=} sign (e.g., \texttt{shop=supermarket})
  \item Do \textbf{not} include bare keys such as \texttt{shop} or \texttt{amenity}
  \item Do \textbf{not} include explanations or additional text
  \item Do \textbf{not} include Python code markers
  \item Do \textbf{not} use tags unrelated to business activities (e.g., \texttt{landuse=forest})
  \item Output \textbf{only} the Python list
\end{itemize}

\textbf{OSM Tags:}

\end{tcolorbox}
\end{figure}

\subsection{Output Prompts}

We have two output prompts. The \texttt{Text} prompts instructs MLLM to return a single token answer. It is used in the baseline configuration. If an MLLM cannot identify the class it returns the class as \texttt{UNK}. The other configuration, used for explanations, is \texttt{JSON} output prompt. The JSON output starts with the explanation less than 50 words and followed by the classification decision.

\begin{figure}[!h]
\centering
\begin{tcolorbox}[
  colback=blue!4,
  colframe=blue!70!black,
  title=\textbf{Text Output Prompt},
  fonttitle=\bfseries,
  sharp corners,
  boxrule=0.8pt,
  width=0.95\columnwidth
]

You will return only the \textbf{SECTOR CODE}. \\ 
If you are not sure about the sector code,\\return \texttt{"UNK"} as a default value.

\textbf{Example}
\texttt{A}\\
\textbf{SINGLE TOKEN RESPONSE ONLY}
\end{tcolorbox}
\end{figure}

\begin{figure}[!h]
\centering
\begin{tcolorbox}[
  colback=blue!4,
  colframe=blue!70!black,
  title=\textbf{JSON Output Prompt},
  fonttitle=\bfseries,
  sharp corners,
  boxrule=0.8pt,
  width=0.95\columnwidth
]
You will return a JSON output including the Sector and Explanation. 
Explanation should be a short description, less than 50 words, 
of why you chose this sector code.

\begin{lstlisting}[breaklines=true,basicstyle=\ttfamily\footnotesize]
{
  "EXPLANATION": "This belongs to Category A because ...",
  "LLM_RESPONSE": "A"
}
\end{lstlisting}
\textbf{DO NOT PRINT ANYTHING OTHER THAN JSON RESPONSE}
\end{tcolorbox}
\end{figure}

\begin{figure*}[!h]
\centering
\begin{tcolorbox}[
  colback=blue!4,
  colframe=blue!70!black,
  title=\textbf{Zero-Shot NACE Classification Prompt},
  fonttitle=\bfseries,
  sharp corners,
  boxrule=0.8pt
]

\textbf{Role}\\
You are an assistant designed to identify \textit{economic activities} from heterogeneous geospatial and textual resources.

\textbf{Inputs}
\begin{itemize}
  \item \textbf{Images}: OpenStreetMap (OSM), Satellite imagery
  \item \textbf{Textual}: Wikidata, Wikipedia, Website
  \item \textbf{Entity name}
\end{itemize}

\textbf{Visual Analysis (Images)}
Identify relevant geospatial features, including but not limited to:
\begin{itemize}
  \item Buildings
  \item Terrain
  \item Streets
\end{itemize}

\textbf{Contextual Analysis (Text)}
Extract economic context such as:
\begin{itemize}
  \item Products and services
  \item Activities
  \item Business type
  \item Industry
\end{itemize}

\textbf{Task}\\
Based on the extracted attributes and the entity name, predict the corresponding
\textbf{NACE Rev.2 economic activity sector code}.

\begin{center}
\fbox{\texttt{\{NACE\_CONTEXT\}}}
\end{center}

\textbf{Available Resources}
\begin{itemize}
  \item \texttt{osm}: OSM image
  \item \texttt{satellite}: Satellite image
  \item \texttt{source}: Wikidata / Wikipedia / Website
\end{itemize}

If no external resources are provided, rely solely on the entity name.

\textbf{Output Format}
\begin{center}
\fbox{\texttt{\{OUTPUT\_FORMAT\}}}
\end{center}

\end{tcolorbox}
\end{figure*}

\subsection{Zero Shot}

In the Zero-Shot classification prompt, we define instructions for the input types to guide MLLMs feature extraction process. Then we provide NACE\_CONTEXT. This context can be \texttt{Simple} (NACE Codes and Titles) and \texttt{Extended} (NACE Codes, Titles and AI generated summaries from official guidelines).

In the classification prompt, we define the set of available inputs and as we test the models without any inputs, we instruct the model to use the name if no input is provided. Then, we provide the output prompt depending on the experiment setup. Finally, we give the context based on the input configuration.

\begin{figure*}[!h]
\centering
\begin{tcolorbox}[
  colback=blue!4,
  colframe=blue!70!black,
  title=\textbf{Clue Extraction Agent Shared Instructions},
  fonttitle=\bfseries,
  sharp corners,
  boxrule=0.8pt,
  width=\textwidth
]

You are an agent tasked with extracting \textit{explicit economic activity clues}
from a single information source.

\textbf{General Rules}
\begin{itemize}
  \item Only extract activities with \textbf{direct textual or visual evidence}.
  \item The provided keyword list defines all valid economic activity categories.
  \item Match only exact keywords or clear synonyms.
  \item Do \textbf{not} infer, guess, or generalize beyond the source.
  \item When mentioning an activity, wrap it in \texttt{[ ]} exactly as in the keyword list.
  \item For every activity, cite the exact supporting feature, tag, phrase, or entity.
  \item If no activity is present, output exactly:       \texttt{"No economic activity clues found."}
  \item Output language must be English.
  \item Maximum output length: 512 tokens.
\end{itemize}

\textbf{Output Format}

Economic activity clues:\\
- [keyword] supporting evidence from the source

\end{tcolorbox}
\end{figure*}

\begin{figure*}[!htpb]
\centering
\begin{tcolorbox}[
  colback=blue!4,
  colframe=blue!70!black,
  title=\textbf{Multi-Turn Classification Prompt},
  fonttitle=\bfseries,
  sharp corners,
  boxrule=0.8pt,
  width=\textwidth
]

\textbf{Role}\\
You are an assistant designed to identify \textit{economic activities} from multiple, incremental information sources.

\vspace{0.5em}
\textbf{Inputs}\\
You may be provided with clues from the following sources:\\
Wikidata, Wikipedia, Websites, OpenStreetMap (OSM) images, Satellite images

\textbf{Task}\\
Based on the provided clues and the entity name, identify the corresponding
\textbf{NACE economic activity sector code}.

\begin{center}
\fbox{\texttt{\{NACE\_CONTEXT\}}}
\end{center}

Note that you may not be given all of the clues.  
If no clues are provided, rely solely on the entity name.

\textbf{Output Format}
\begin{center}
\fbox{\texttt{\{OUTPUT\_FORMAT\}}}
\end{center}

\end{tcolorbox}
\end{figure*}

\subsection{Multi Turn}

In our multi-turn pipeline, we have intermediate-level agents for each input type (Satellite, OSM, Wikidata, Wikipedia, and Website). Each processor agent prompt, contains several instructions for data processing and sets the generation limit to 512 tokens. MLLMs are instructed to generate \textbf{No Economic Activity Found} in response if they cannot retrieve evidence from an input. For each processor, we give NACE Keywords defined in Table~\ref{tab:appendix_nace_content}. These keywords are expected in the output for easier grouping of the free-form text.  We provided a sample prompt containing shared instructions here.

\begin{figure}[]
\centering
\begin{tcolorbox}[
  colback=blue!4,
  colframe=blue!70!black,
  title=\textbf{Ablation Instructions },
  fonttitle=\bfseries,
  sharp corners,
  boxrule=0.8pt,
  width=\textwidth
]
Classify the company into one industry sector.

You are given codes and titles.

Respond with EXACTLY ONE UPPERCASE LETTER. 

Do NOT include spaces, newlines, punctuation, or any other text.

If unsure, pick the single best letter based on the company’s \textbf{primary revenue-generating activity}.

VALID LETTERS: [Available options]
\end{tcolorbox}
\end{figure}
\begin{figure}[]
\centering
\begin{tcolorbox}[
  colback=blue!4,
  colframe=blue!70!black,
  title=\textbf{ExioNAICS Prompt},
  fonttitle=\bfseries,
  sharp corners,
  boxrule=0.8pt,
  width=\columnwidth
]
\begin{center}
\fbox{\texttt{\{ABLATION\_INSTRUCTIONS\}}}
\end{center}
Choices (A–T): Title\\
A: Agriculture, Forestry, Fishing and Hunting\\
B: Mining, Quarrying, and Oil and Gas Extraction\\
C: Utilities\\
D: Construction\\
E: Manufacturing\\
F: Wholesale Trade\\
G: Retail Trade\\
H: Transportation and Warehousing\\
I: Information\\
J: Finance and Insurance\\
K: Real Estate and Rental and Leasing\\
L: Professional, Scientific, and Technical Services\\
M: Management of Companies and Enterprises\\
N: Administrative and Support and Waste Management and Remediation Services\\
O: Educational Services\\
P: Health Care and Social Assistance\\
Q: Arts, Entertainment, and Recreation\\
R: Accommodation and Food Services\\
S: Other Services (except Public Administration)\\
T: Public Administration\\
\end{tcolorbox}
\end{figure}

\begin{figure}[!h]
\centering
\begin{tcolorbox}[
  colback=blue!4,
  colframe=blue!70!black,
  title=\textbf{Company Websites Prompt },
  fonttitle=\bfseries,
  sharp corners,
  boxrule=0.8pt,
  width=\columnwidth
]
\begin{center}
\fbox{\texttt{\{ABLATION\_INSTRUCTIONS\}}}
\end{center}

Choices (A–M): Title:\\
A: Commercial Services \& Supplies\\
B: Healthcare\\
C: Materials\\
D: Financials\\
E: Energy \& Utilities\\
F: Professional Services\\
G: Corporate Services\\
H: Media, Marketing \& Sales\\
I: Information Technology\\
J: Consumer Discretionary\\
K: Industrials\\
L: Transportation \& Logistics\\
M: Consumer Staples\\
\end{tcolorbox}
\end{figure}

Clues are appended to the Multi-Turn classification prompt after construction. Final decision-making agent prompt resembles the single-stage Zero-Shot classification prompt. It replaces the entries with the text clues.

\subsection{Ablation Prompts}
Based on our zero-shot template, we designed prompts for text-only benchmarks ExioNAICS \cite{guo_group_2025} and Company Websites \cite{rizinski_company_2023}. For the few-shot examples, we randomly selected with a fixed seed one example per class from the training set. The prompt structures are identical for each task. It starts with a set of instructions followed by available categories and choices.

\end{document}